\documentclass[journal]{IEEEtran}

\usepackage{times}
\usepackage{epsfig}
\usepackage{graphicx}
\usepackage{amssymb}
\usepackage[tight]{subfigure}
\usepackage{caption}
\usepackage{subfig}
\usepackage{multirow}
\usepackage{comment}
\usepackage{caption}
\usepackage[shortlabels]{enumitem}
\usepackage{array}
\usepackage{threeparttable}
\usepackage{amsmath}
\usepackage{url}

\makeatletter
\def\footnoterule{\relax%
 \kern-5pt
 \hbox to \columnwidth{\hfill\vrule width \columnwidth height 0.6pt\hfill}
 \kern4.6pt}
\makeatother

\newcolumntype{P}[1]{>{\centering\arraybackslash}p{#1}}
\newcolumntype{M}[1]{>{\centering\arraybackslash}m{#1}}

\usepackage{algorithm}
\usepackage[noend]{algpseudocode}
\makeatletter
\def\BState{\State\hskip-\ALG@thistlm}
\makeatother

\setlist[enumerate]{nosep}
\setlist[itemize]{nosep}

\usepackage{multirow}
\usepackage{hhline}%
\usepackage{booktabs}
\usepackage{tabularx}

\usepackage{cite}

\DeclareMathOperator*{\argmin}{arg\,min}

\ifCLASSINFOpdf
\else
\fi

\begin{document}
%
\title{Face Clustering: Representation and Pairwise Constraints}
%

\author{Yichun~Shi,~\IEEEmembership{Student Member,~IEEE,}
     Charles~Otto,~\IEEEmembership{Member,~IEEE,}
     and Anil K. Jain,~\IEEEmembership{Fellow,~IEEE}

\thanks{Y. Shi and A. K. Jain are with the Department of Computer Science and Engineering, Michigan State University, East Lansing, MI, 48824. E-mail: shiyichu@msu.edu, jain@cse.msu.edu}
\thanks{C. Otto is with Noblis, Reston, VA. E-mail: ottochar@gmail.com}
}

%
%

\markboth{}
{Shell \MakeLowercase{\textit{et al.}}: Bare Demo of IEEEtran.cls for IEEE Journals}
%



\maketitle

\begin{abstract}
Clustering face images according to their latent identity has two important applications: (i) grouping a collection of face images when no external labels are associated with images, and (ii) indexing for efficient large scale face retrieval. The clustering problem is composed of two key parts: representation and similarity metric for face images, and choice of the partition algorithm. We first propose a representation based on ResNet, which has been shown to perform very well in image classification problems. Given this representation, we design a clustering algorithm, Conditional Pairwise Clustering (ConPaC), which directly estimates the adjacency matrix only based on the similarities between face images. This allows a dynamic selection of number of clusters and retains pairwise similarities between faces. ConPaC formulates the clustering problem as a Conditional Random Field (CRF) model and uses Loopy Belief Propagation to find an approximate solution for maximizing the posterior probability of the adjacency matrix. Experimental results on two benchmark face datasets (LFW and IJB-B) show that ConPaC outperforms well known clustering algorithms such as \textit{k}-means, spectral clustering and approximate Rank-order. Additionally, our algorithm can naturally incorporate pairwise constraints to work in a semi-supervised way that leads to improved clustering performance. We also propose an \textit{k}-NN variant of ConPaC, which has a linear time complexity given a \textit{k}-NN graph, suitable for large datasets.
\end{abstract}


%
\IEEEpeerreviewmaketitle
\section{Introduction}

\IEEEPARstart{C}{ameras} are everywhere, embedded in billions of smart phones and hundreds of millions of surveillance systems. Surveillance cameras, in particular, are a popular security mechanism employed by government agencies and businesses alike. This has resulted in the capture of suspects based on their facial images in high profile cases such as the 2013 Boston Marathon Bombing~\cite{klontz2013case}. But, getting to the point of locating suspects' facial images typically requires manual processing of large volumes of images and videos of an event. The need for automatic processing of still images and videos to assist in forensic investigations has motivated prior works on clustering large collections of faces by identity~\cite{otto2017clustering}. In~\cite{nech2017level}, Nech et al. also used face clustering to help label face images and compiled MF2 face dataset.

In surveillance applications, the quality of available face images is typically quite low compared to face images in some of the public domain datasets such as the Labeled Faces in the Wild (LFW)~\cite{LFWTech}. The IARPA Janus project is pushing the boundaries of unconstrained face recognition and has released a dataset, NIST IJB-B~\cite{ijb-b2017}\footnote{There is also an earlier version with smaller number of images, called IJB-A released in 2015~\cite{klare2015pushing}}, where many of the faces cannot be detected by off-the-shelf face detectors~\cite{viola2004robust}. The face recognition problem posed by the Janus benchmark may therefore be closer to that encountered in forensic applications. We attempt to handle this more difficult category of faces by: (i) improving the face representation (through the use of large training sets, and new deep network architectures), (ii) developing an effective face clustering algorithm to automatically group faces in images and videos, and (iii) incorporating user feedback during the clustering process via a semi-supervised extension of the proposed clustering algorithm.

To develop a representation for face clustering, we leverage two public domain datasets: CASIA-Webface~\cite{yi2014learning} and VGG-Face~\cite{parkhi2015deep}. 
In terms of network architecture, we adopt deep residual networks, which have resulted in better performance than VGG-architecture on the ImageNet benchmark~\cite{kaminghe2016residual}, and improved results over the architecture proposed in~\cite{yi2014learning} on the BLUFR protocol~\cite{BLUFR}.

\begin{figure}[t]
\center
\includegraphics[width=\linewidth]{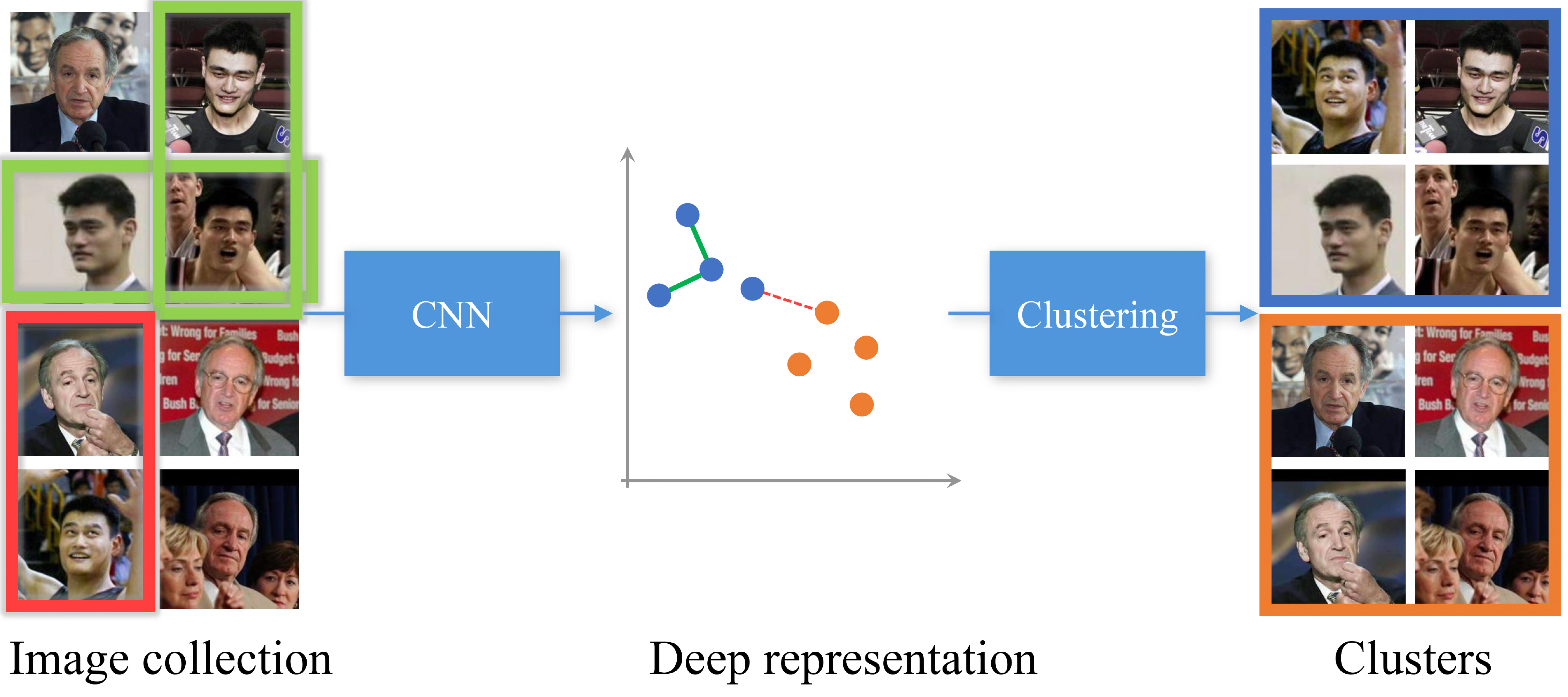}
   \caption{Face clustering workflow. A deep neural network is trained to generate the representations for aligned face images. Given the representation and a similarity measure, goal of clustering is to group these unlabeled face images according to their identity. In the semi-supervised scenario, pairwise constraints are provided along with the unlabeled data. The red line here indicates a cannot-link pair and green lines indicate must-link pairs.}
\label{fig:constraint}
\end{figure}

Given a representation, we propose a face clustering method, called \textit{Conditional Pairwise Clustering (ConPaC)} to group the face collection according to their hidden class (subject identity) using the pairwise similarities between face images. No assumption about the data, including the true number of identities (clusters), is used; only a threshold on similarity is specified to balance between the precision and recall rate. Instead of learning new similarity measures or representations and feeding them to a standard partitional or hierarchical clustering method, Conditional Pairwise Clustering directly treats the adjacencies between all pairs of faces as the variables to predict and look for a solution that maximizes the joint posterior probability of these variables given their corresponding pairwise similarities. To model this conditional distribution, we propose a triplet consistency constraint which reveals such a dependency between the output variables that a valid adjacency matrix must be transitive to represent a partitional clustering. That means any two adjacent points should share exactly the same adjacent neighbors. The proposed model can dynamically determine the number of clusters, and also retain the similarity information. In particular, we model the problem as a Conditional Random Field (CRF) and employ Loopy Belief Propagation to arrive at a valid adjacency matrix. This model is easily extended to the semi-supervised clustering by accepting a set of pairwise constraints (either must-link, or cannot-link assignments) on the similarity matrix.

We perceive the following contributions in this work: (i) We propose a clustering algorithm (ConPaC) based on direct estimation of an adjacency matrix derived from pairwise similarities between faces using the learned representation from a Deep Residual Network; (ii) We evaluate the proposed method on two unconstrained face datasets: LFW and IJB-B; (iii) We show that the proposed method can be naturally applied to semi-supervised face clustering scenario; (v) We propose an approximate \textit{k}-NN variant of the algorithm for efficient clustering of millions of face images. 


\section{Background}

\subsection{Face Representation}
Face images have been traditionally represented by appearance models or local descriptors~\cite{turk1991face}~\cite{cootes2001active}~\cite{ahonen2006face}~\cite{wright2009robust}. But as Deep Neural Networks (DNN) have shown their great potential in solving computer vision problems due to its representation learning ability~\cite{krizhevsky2012imagenet}\cite{ren2015faster}\cite{hsu2017cnn}, a number of DNN based methods have been proposed for face representation and recognition. The DeepFace~\cite{taigman2014deepface} method trained a CNN on a dataset of four million facial images belonging to more than 4,000 identities. The training is based on minimizing classification error and the output of the last hidden layer taken as the face representation. DeepFace significantly surpassed the traditional methods in face recognition, especially for unconstrained face images. Sun et al. extended the work of DeepFace in their DeepId series~\cite{sun2014deepid}~\cite{sun2014deepid2}~\cite{sun2015deeply}~\cite{sun2015deepid3}. They proposed to use multiple CNNs with joint Bayesian framework~\cite{chen2012bayesian} and added supervision to early convolutional layers. Schroff et al.~\cite{schroff2015facenet}, in their FaceNet work, abandoned the classification layer and instead introduced the triplet loss to directly learn an embedding space where feature vectors of different identities could be separated with Euclidean distance.

\subsection{Face Clustering}

Cluster analysis is an important topic widely studied in pattern recognition, statistics and machine learning~\cite{jain2010data}. It is useful for exploratory analysis by a preliminary grouping of a collection of unlabeled data. Due to potentially large and unknown number of identities in many large scale face collections, it is useful to tag the face images with the labels obtained from clustering. Otto et al.~\cite{otto2017clustering} provided a brief review of face clustering. Most of the previous studies~\cite{ho2003clustering}~\cite{cui2007easyalbum}~\cite{tian2007face}~\cite{elhamifar2009sparse}~\cite{zhu2011rank}~\cite{vidal2014low} focused on learning a good similarity matrix or robust representations from non-discriminative low-level features and then partitioned the dataset with standard clustering algorithms such as spectral clustering. However, in practice, high-level features generated by deep neural networks today can give quite robust representation, and hence similarity, even with simple metric functions. Furthermore, many supervised metric or representation learning methods have been proposed, which are shown to be able to enhance the deep representations and have good generalizability~\cite{chen2012bayesian}~\cite{sankaranarayanan2016triplet}. Therefore, as we will show in Section~\ref{sec:motivation}, similarity learning is relatively simple in practical face clustering problems; instead the partitioning algorithm plays a more important role. 

Otto et al.~\cite{ottoefficient}~\cite{otto2017clustering}, based on the work of~\cite{zhu2011rank}, made use of the assumption that homogeneous face images (images that belong to the same identity) usually have similar nearest neighbors and proposed the approximate Rank-order distance metric. They showed that by linking all the image pairs within a certain distance, they can achieve good clustering performance on challenging unconstrained face datasets.

\subsection{Semi-supervised Clustering}

Given the difficult nature of data clustering (choice of representation, similarity measure and number of clusters), one approach to improve clustering performance is to incorporate side-information. One common form of side-information is pairwise constraints, indicating that a pair of data points either must be placed in the same cluster (a ``must-link" constraint), or they cannot be placed in the same cluster (a ``cannot-link" constraint), as shown in Figure~\ref{fig:constraint}. Wagstaff et al.~\cite{wagstaff2001constrained} first incorporated the pairwise constraints into \textit{k}-means algorithm by forcing the cluster assignments to satisfy the constraints and showed that user-specified constraints could help to improve clustering results. Xing et al.~\cite{xing2002distance} proposed to learn a Mahalanobis distance metric from the given constraints before applying \textit{k}-means. Basu et al.~\cite{basu2004probabilistic} designed a probabilistic model for semi-supervised clustering with Hidden Markov Random Fields (HMRFs) and used EM algorithm to optimize the parameters. Research has also been conducted on incorporating pairwise constraints into hierarchical clustering~\cite{davidson2005agglomerative} and spectral clustering~\cite{lu2008constrained}~\cite{wang2010flexible}. For a review of semi-supervised clustering, readers are referred to~\cite{basu2008constrained}.


\begin{figure*}[t]
\center
\includegraphics[width=\linewidth]{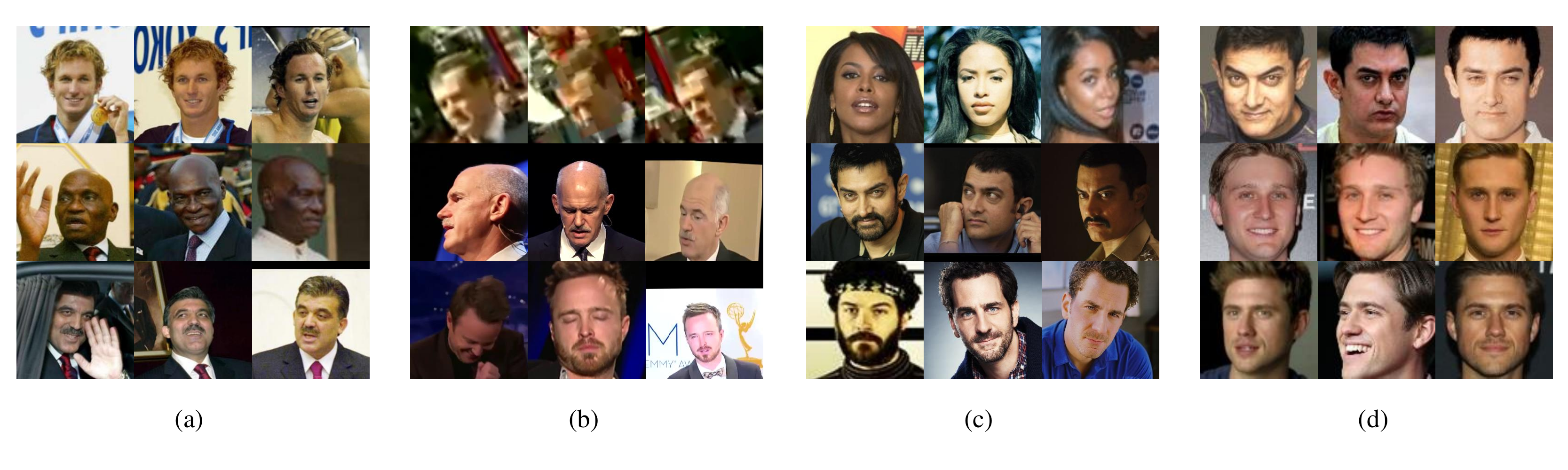}
   \caption{Example face images from (a) LFW, (b) IJB-B, (c) CASIA-webface, and (d) VGG datasets.}
\label{fig:datasets}
\end{figure*}

\begin{figure}[t]
\center
\includegraphics[width=0.4\linewidth]{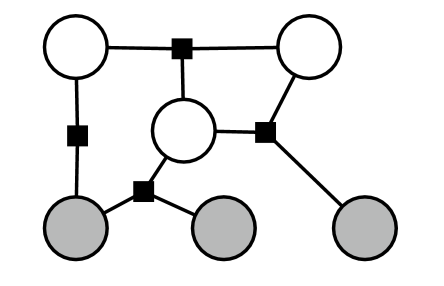}
   \caption{An example factor graph of a general CRF model. Here, the filled nodes are input nodes and the white nodes are output nodes. Each factor (square) represents a potential function on a clique, encoding the contraints between nodes. There are constraints either between input and output or between output nodes. The figure is taken from~\cite{sutton2006introduction}.}
\label{fig:general_crf}
\end{figure}

\subsection{Conditional Random Fields (CRFs)}
\label{sec:crf}
Conditional Random Fields (CRFs) are a type of undirected probabilistic graphical models first proposed by Lafferty et al. for predicting labels of sequential data~\cite{lafferty2001conditional} and later introduced to computer vision to model images~\cite{he2004multiscale}~\cite{hoiem2008putting}~\cite{koltun2011efficient}. The difference between CRFs and traditional Hidden Marcov Models (HMMs)~\cite{rabiner1989tutorial} lies in that CRF is a discriminant model which directly models the conditional distribution $p(Y|X)$ rather than joint distribution $p(Y,X)$ and predicts the labels $Y$ by maximizing the posterior probability. Generally, a CRF can be formulated as:
\begin{equation}
p(Y|X)=\frac{1}{Z}\prod_{C_p\in C}\prod_{\psi_c\in C_p}\psi_c(X_c,Y_c;\theta_p),
\end{equation}
where $Z$ is the normalization factor, $C=\{C_1,C_2,...,C_P\}$ is the set of all cliques in the graph, $\psi_c$ is a potential function defined on the variables $(X_c,Y_c)$ in clique $C_p$, and $\theta_p$ is a set of parameters of the model~\cite{sutton2006introduction}. We can represent undirected graphical models by a factor graph, where a factor node is there for each potential function and connects to every node in its clique, as shown in Figure~\ref{fig:general_crf}. Usually, there are two types of potential functions in CRFs: (1) association potential that equals the local conditional distribution over observations $p(Y_c|X_c)$ and (2) interaction potential that encodes the dependencies between different output variables. Although both of them are originally defined as Gibbs distributions on features in~\cite{lafferty2001conditional}, association potentials are often substituted by supervised discriminant classifiers such as neural networks~\cite{he2004multiscale}~\cite{chen2014semantic}.

As for inference on the CRFs, any method for undirected graphical models can be applied, and one of these methods is Belief Propagation (BP)~\cite{pearl1988probabilistic}. There are two types of BP algorithms: sum-product and max-sum. They are exact inference methods, respectively, for finding marginal probability and maximizing posterior probability on tree-like graphical models. But because they only involve local message updates, they can also be applied to graphs with loops, resulting in Loopy Belief Propagation. Although Loopy Belief Propagation is not guaranteed to converge, it has achieved success in a variety of domains~\cite{freeman2000learning}~\cite{frey1998revolution}~\cite{yanover2002approximate}. It has also been shown that the result of Loopy Belief Propagation corresponds to the stationary point of Bethe free energy and that it is related to variational methods~\cite{yedidia2005constructing}. Readers are referred to~\cite{sutton2012introduction} for further information on CRFs and Loopy Belief Propagation.

\section{Face Datasets}
We leverage the CASIA-Webface~\cite{yi2014learning}, and VGG-Face~\cite{parkhi2015deep}  datasets to train networks for learning the representation to be used for clustering. We then evaluate the performance of our clustering algorithm on two benchmarks datasets, LFW~\cite{LFWTech}, and IARPA Janus Benchmark-B (IJB-B). Some example images from these datasets are shown in Figure~\ref{fig:datasets}. As stated in~\cite{yi2014learning} and~\cite{parkhi2015deep}, there is no overlap of identity between LFW and VGG-Face or LFW and CASIA-Webface. Similarly, IJB-B does not include overlapping identities with VGG-Face or CASIA-Webface~\cite{ijb-b2017}.

\subsection{LFW}
The Labeled Faces in Wild (LFW)~\cite{LFWTech} contains $13,233$ face images of $5,749$ individuals; of those $5,749$ individuals, $4,069$ have only one face image each. The dataset was constructed by searching for images of celebrities and public figures, and retaining only those images for which an automatically detectable face via the off-the-shelf face detectors~\cite{viola2004robust} was present. As a result, facial pose variations in LFW are limited.

\subsection{IJB-B}
The IJB-B dataset~\cite{ijb-b2017} is composed of $7$ different clustering experiments, with increasing number of subjects. These experiments, respectively, involve $32$, $64$, $128$, $256$, $512$, $1,024$, $1,845$ subjects with total of $1,026$, $2,080$, $5,224$, $9,867$, $18,251$, $36,575$ and $68,195$ images, respectively. Two protocols related to clustering are defined for IJB-B dataset: (i) clustering of detected faces and (ii) face detection + clustering. Since the focus of this work is face clustering, we will use the first protocol and assume faces have already been detected. The faces are aligned following the procedure in~\cite{wang2015face}, using the bounding boxes provided in IJB-B as the starting point for landmark detection. Many images in the IJB-B datasets are in extreme poses or of low quality, making the clustering task more difficult for IJB-B than for LFW. Most of the images in the clustering protocols of IJB-B are from video frames, making it more related to the surveillance application.

\subsection{CASIA-Webface}
The CASIA-webface dataset~\cite{yi2014learning} is a semi-automatically collected face dataset for pushing the development of face recognition systems. It contains $494,414$ images of $10,575$ subjects (mostly celebrities) downloaded from internet. However, we are unable to localize faces in some of the images with the face detector from the Dlib library~\footnote{\url{http://dlib.net}}. So we use a subset of CASIA-Webface with $404,992$ face images of $10,533$ subjects to train our network. This dataset has been popular for training deep networks.


\subsection{VGG-Face}
The VGG-Face dataset was released in 2016 as a set of $2.6$ million URLs with corresponding face detection locations~\cite{parkhi2015deep}. We could acquire only 2.2 million images of the original ~2.6 million listed URLs due to broken links. Example VGG images are shown in Figure~\ref{fig:datasets}(d). Additionally, among the images we were able to download, we identified a number of exact duplicate files. On manual examination, these duplicates typically had mislabeled identities, or were placeholder images served by the image host when the original image was no longer available. After excluding duplicate images, we retained 1.7 million images from the VGG dataset. We combined these images with the CASIA-Webface dataset (and merged overlapping subject identities), to get a total of 2.1M images, of 11,326 distinct subjects.

\begin{figure}[t]{
\center
\includegraphics[width=\columnwidth]{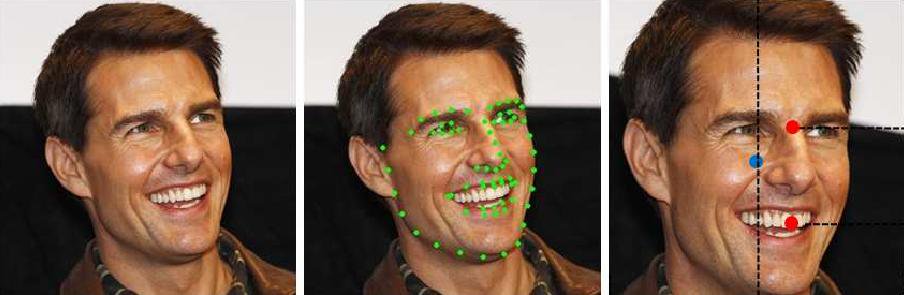}
}
 \caption{An example showing the normalization of face images. We normalize all our images before feeding into the network according to the procedure in~\cite{wang2015face}, where the figure is taken from.}
\label{fig:norm_example}
\end{figure}

\section{Methods}

\subsection{Representation}

He et al.~\cite{kaminghe2016residual} used a ``deep residual network" architecture to achieve competitive results on the ImageNet object recognition dataset. So we directly adapt the architecture for face recognition (leveraging the Torch7 framework\cite{collobert2011torch7} and the implementation of residual networks released by Facebook AI Research fb.resnet.torch\footnote{\url{https://github.com/facebook/fb.resnet.torch}}). We have investigated the 50, and 101-layer architectures outlined in~\cite{kaminghe2016residual}. In terms of data augmentation, we scale our normalized face images following the alignment procedure proposed in~\cite{wang2015face} to $256\times 256$, as shown in Figure~\ref{fig:norm_example}, and randomly crop $224\times 224$ regions during training. We additionally flip images during training, and use the scale and aspect-ratio augmentations from~\cite{szegedy2015going}. As for feature extraction, each image is aligned using the same procedure, then a $2048$-dimension feature vector is extracted from the bottleneck layer\footnote{The last hidden layer. Because a ReLU layer is used as the activation function for the bottleneck layer, all the feature values are non-negative.} with $10$-crop strategy\footnote{Average the features extracted from 10 different sub-crops of size $224\times 224$ (corners + center with/without horizontal flips).} and is normalized into unit $\ell^2$ norm. Because of the low quality of the face images in IJB-B benchmark, many images cannot be aligned using Dlib. In such cases, we crop a square region containing the ground-truth bounding box provided in the protocol. This is allowed in the test 7 of the benchmark, where
the detection task is assumed to be already finished.

\begin{figure*}[t]
\centering
\subfigure[similarity matrix]{
    \includegraphics[width=.22\linewidth]{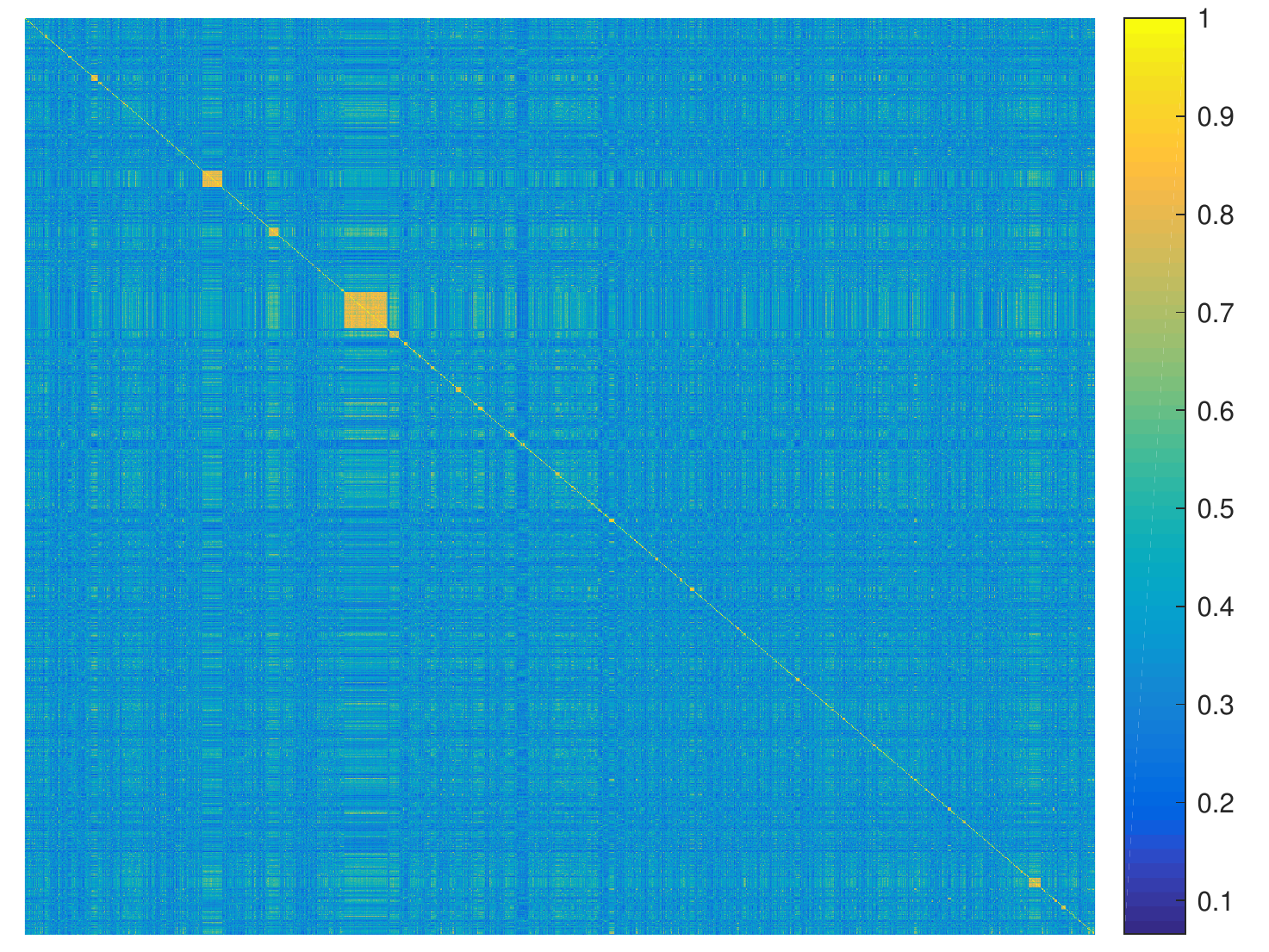}
}
\subfigure[ground-truth]{
    \includegraphics[width=.22\linewidth]{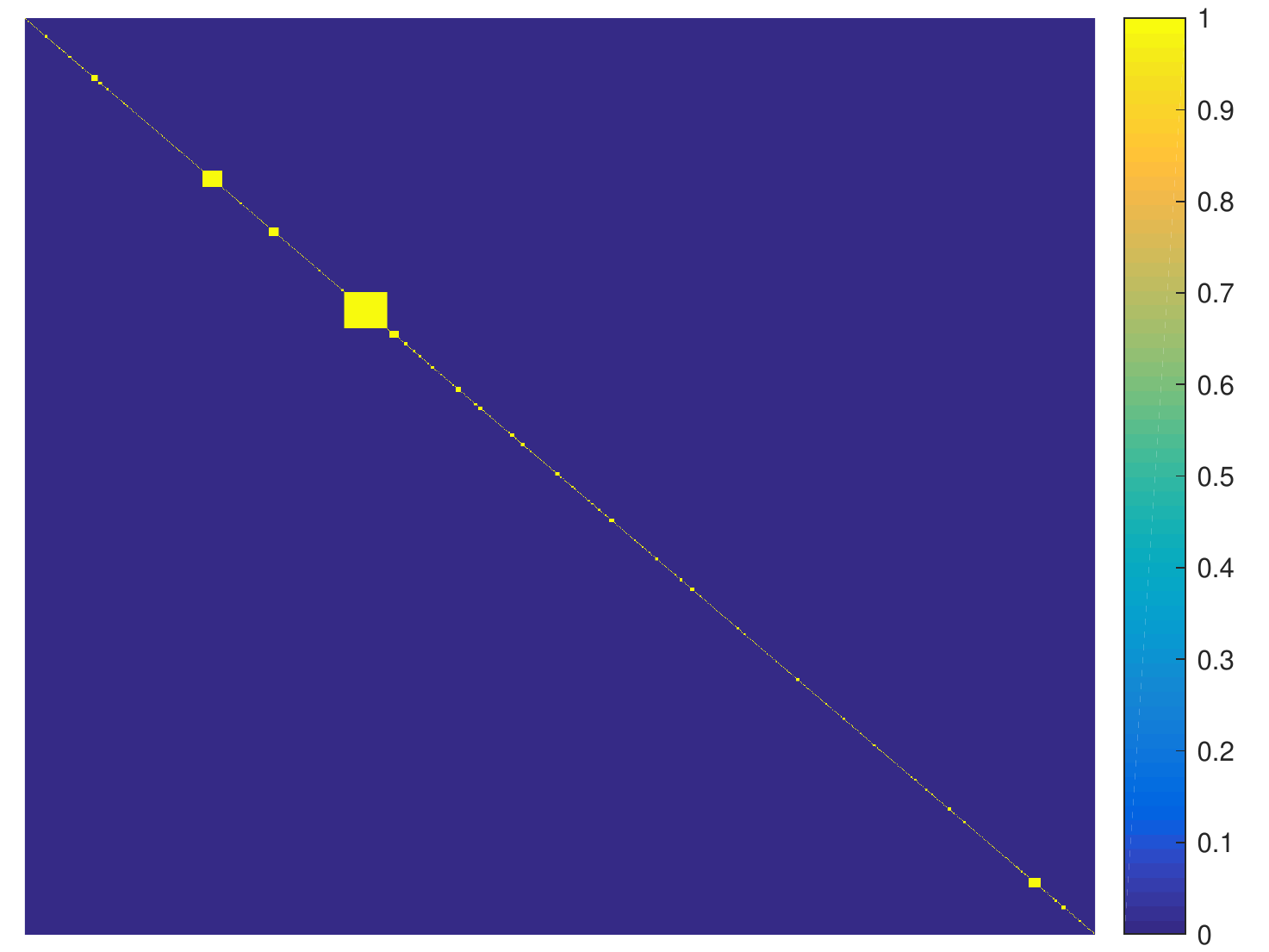}
}
\vspace{-0.25em}
\subfigure[\textit{k}-means ($C=5,749$)]{
    \includegraphics[width=.22\linewidth]{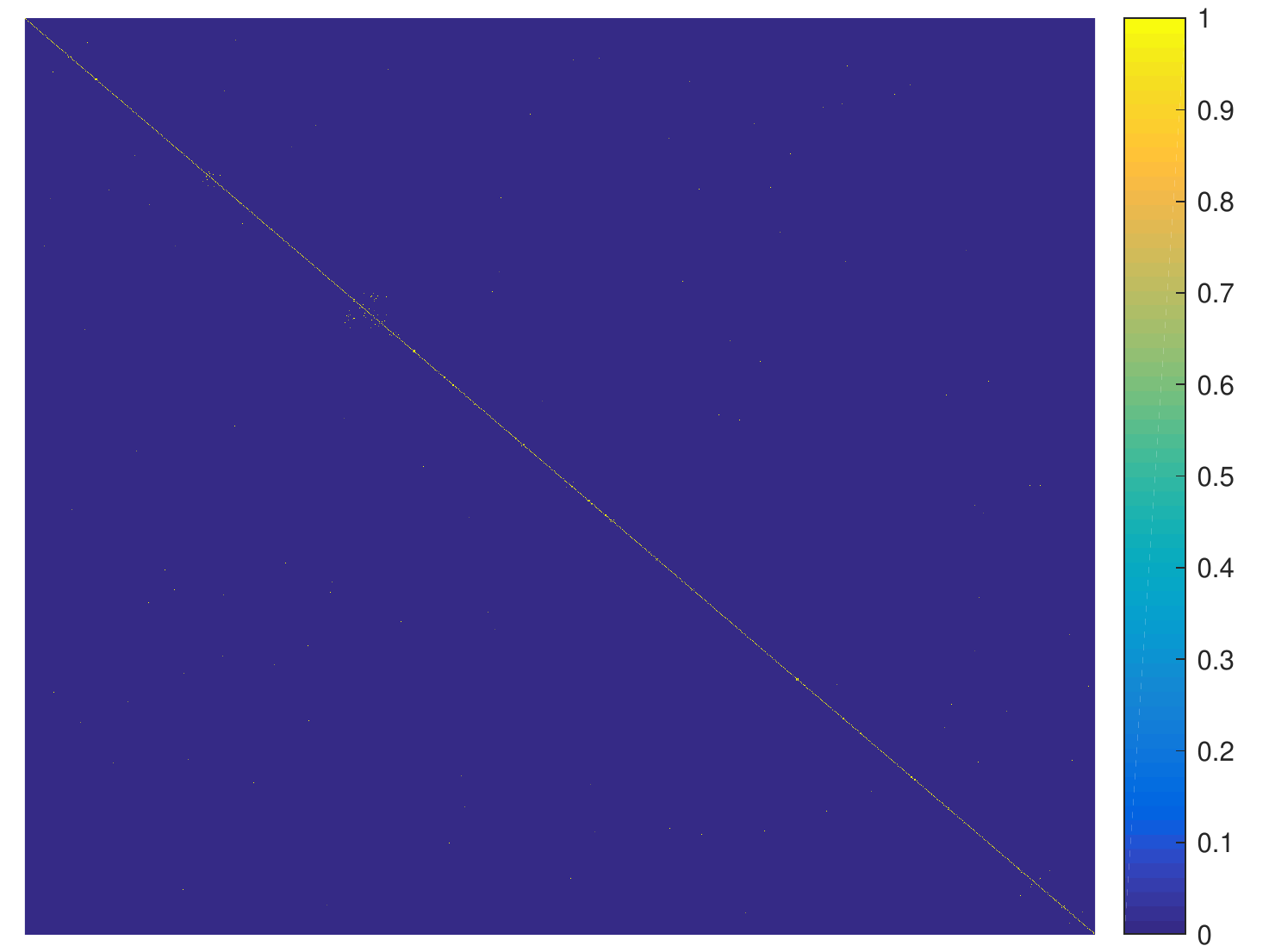}
}
\subfigure[\textit{k}-means ($C=500$)]{
    \includegraphics[width=.22\linewidth]{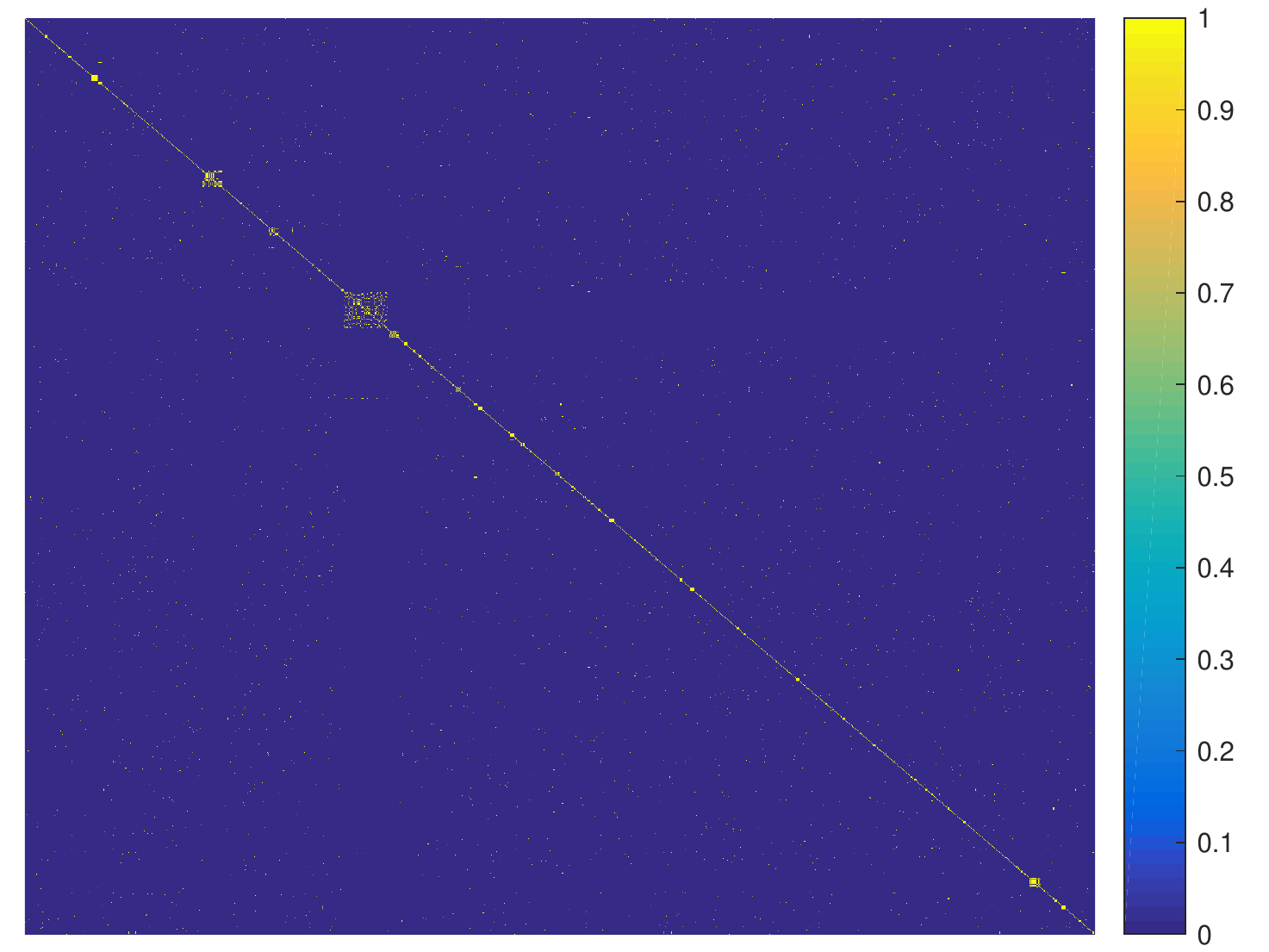}
}\\
\vspace{-0.25em}
\subfigure[spectral clustering ($C=5,749$)]{
    \includegraphics[width=.22\linewidth]{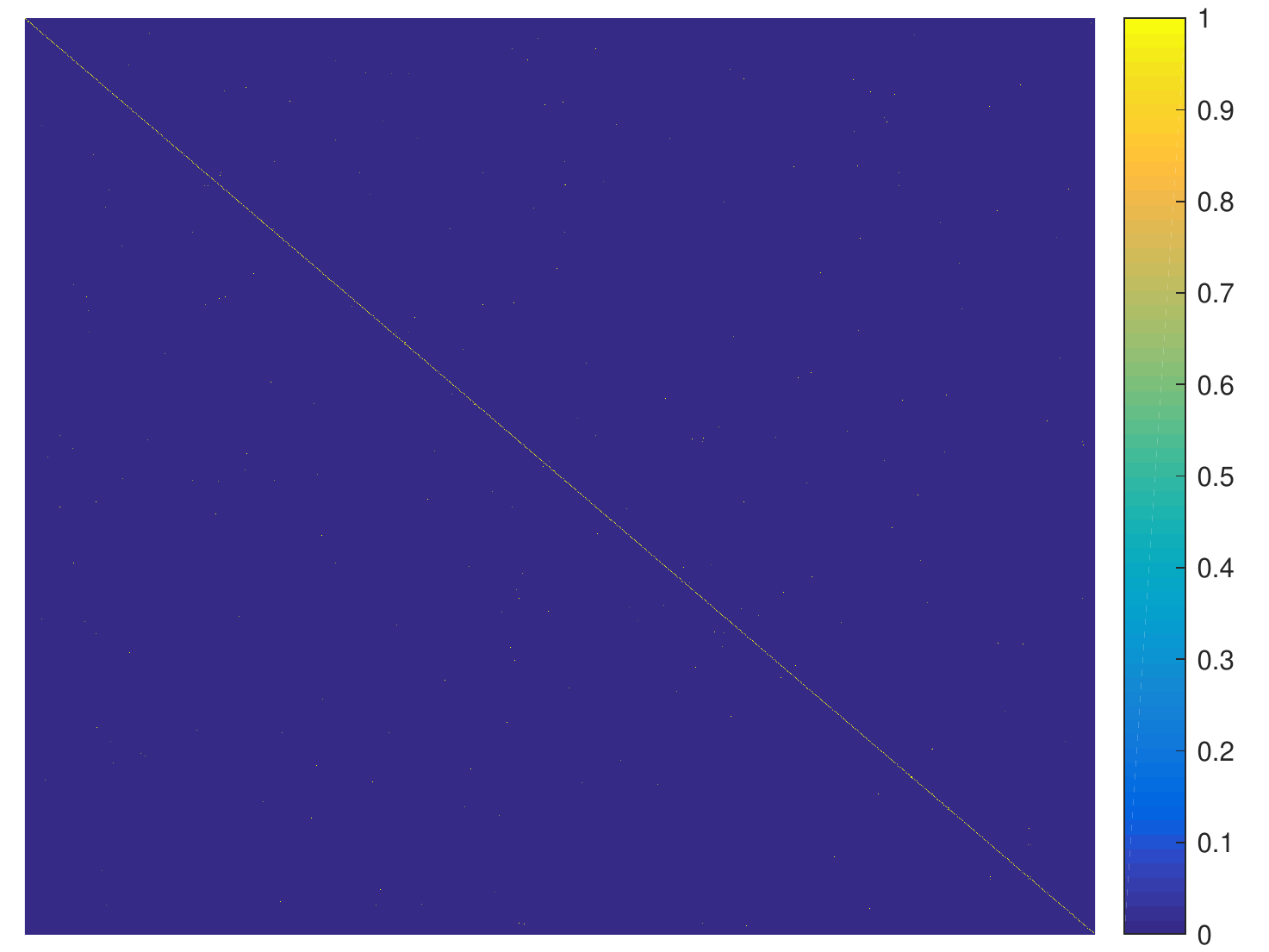}
}
\subfigure[spectral clustering ($C=75$)]{
    \includegraphics[width=.22\linewidth]{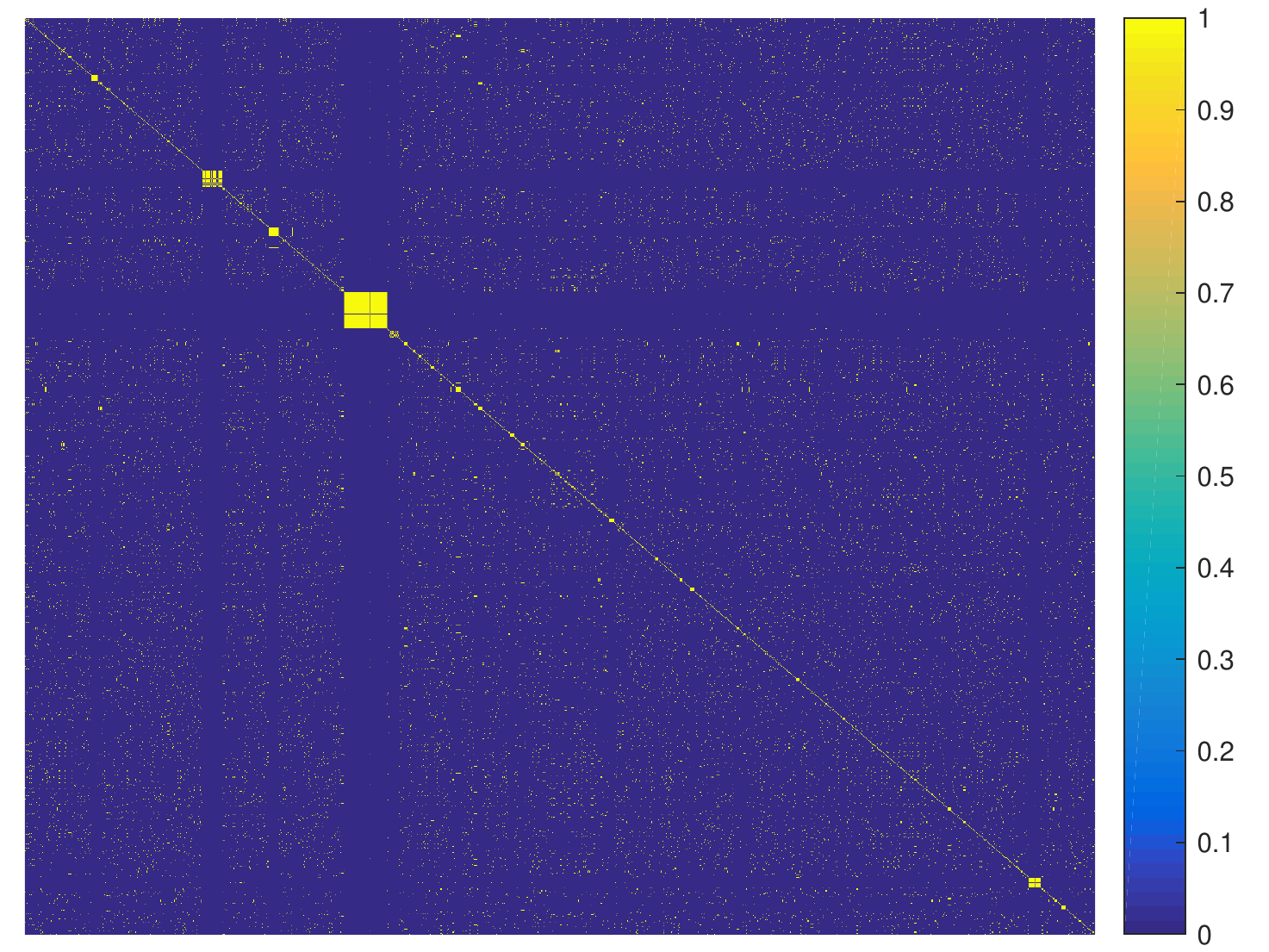}
}
\subfigure[approx. Rank-order]{
    \includegraphics[width=.22\linewidth]{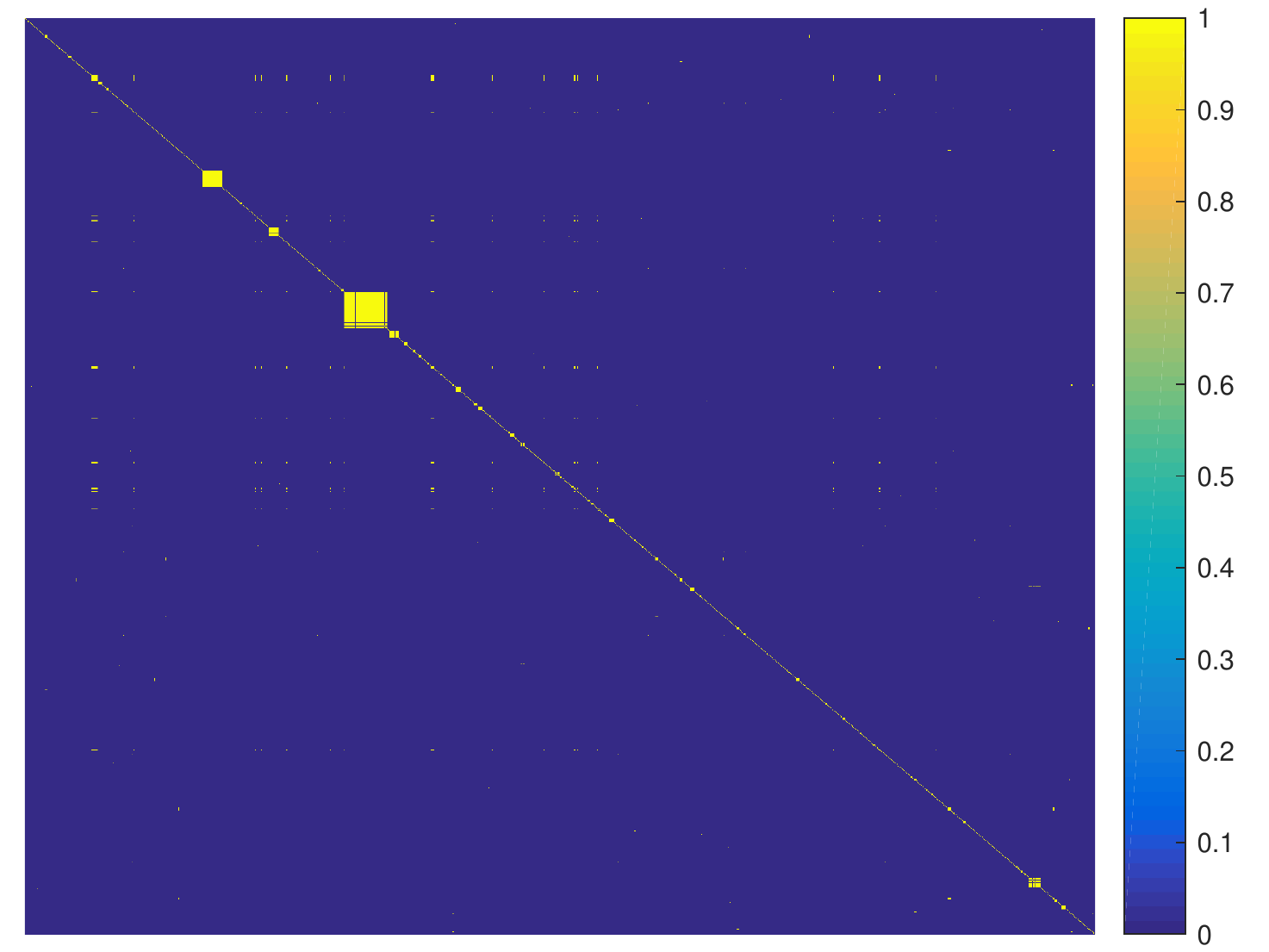}
}
\subfigure[proposed]{
    \includegraphics[width=.22\linewidth]{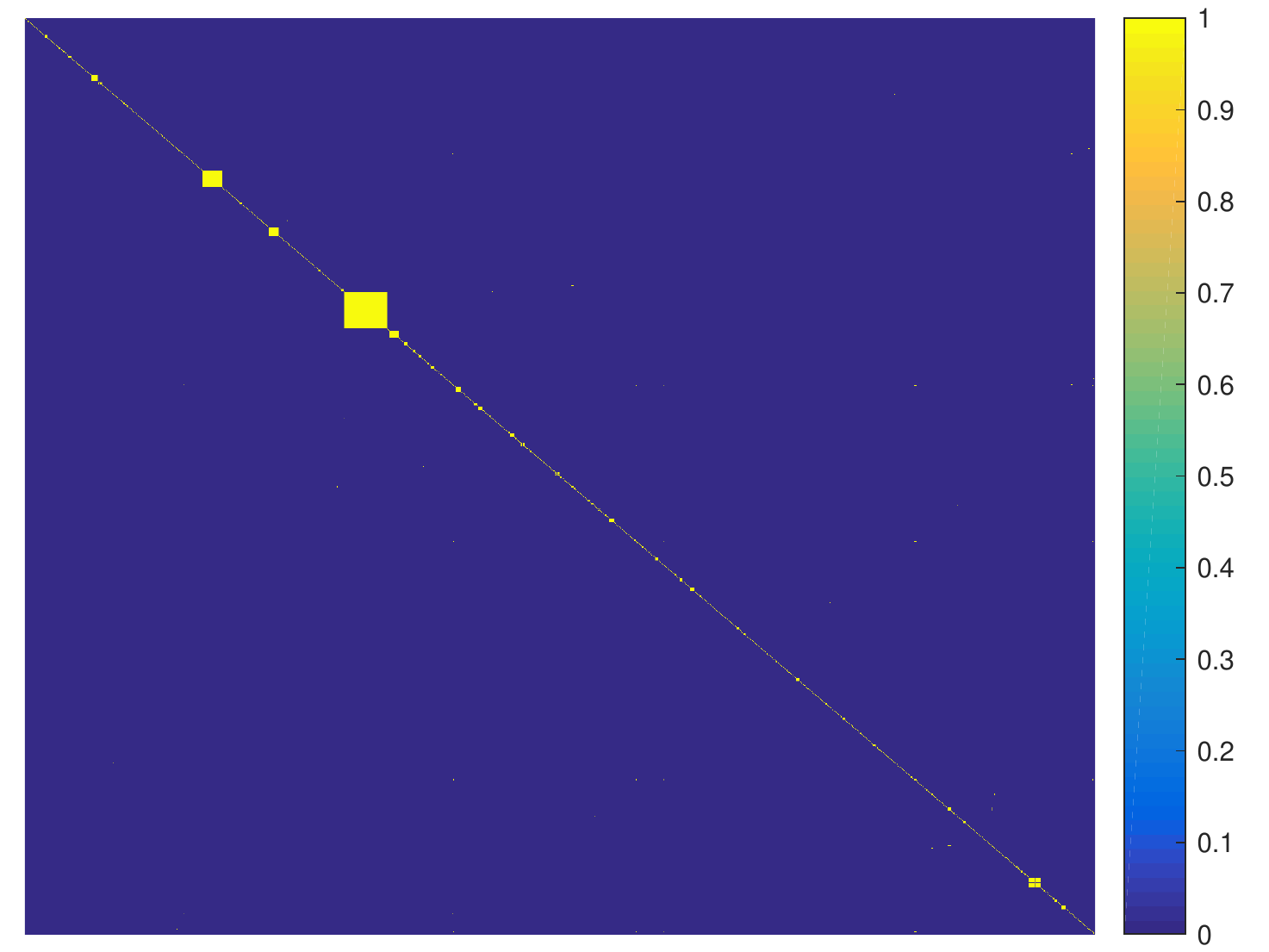}
}\\
\vspace{-0.2em}
\caption{Visualization of the similarity matrix and adjacency matrices. Using the face representations from deep neural networks, similarity matrix (using cosine similarity) is highly consistent with the ground-truth adjacency matrix. However, using the ground-truth number of clusters ($C=5,749$) both \textit{k}-means and spectral clustering fail to retain the similarities and break the big clusters into small groups. Tuning the number $C$ to smaller values give better performance, but makes the clusters less pure. Thus, to make full use of the similarity matrix, we attempt to find a partition by maximizing the consistency between the adjacency matrix and the given pairwise similarities.}
\label{fig:matrix_lfw}
\end{figure*}

\subsection{Motivation}
\label{sec:motivation}

Face clustering, like other clustering problems, attempts to partition data points into a number of groups based on their similarities or structure. However, in real-world problems of unconstrained face clustering, the situation could be quite different such that many popular clustering algorithms are not suitable. On one hand, most clustering algorithms are based on certain distribution assumptions of the data. For example, \textit{k}-means assumes that the data points of different classes are close to the centroids of the classes and spectral clustering~\cite{shi2000normalized}~\cite{ng2001spectral} aims at finding a balanced partition of the dataset. But in fact, the data could be distributed in arbitrary shapes, depending on the representation, and the sizes of different clusters in a large face image collection could be very unbalanced. Besides, most algorithms require the number of clusters as an input parameter, which is usually unknown and quite large in real-world face clustering problems. On the other hand, the rapid development of deep neural networks makes it possible to learn highly robust representations for unconstrained face images. Even with simple metric functions, good pairwise similarities can be acquired, providing reliable evidence on pairwise homogeneity (whether a pair of face images belong to the same identity). 

In Figure~\ref{fig:matrix_lfw}, by visualizing the similarity matrix from ResNet representations, we can see it is highly consistent with the ground-truth adjacency matrix. However, in comparison, the resulting adjacency matrices (by using the ground-truth number of classes) of \textit{k}-means and spectral clustering are not only far from the ground-truth one, but also not similar to the input similarity matrix itself. Thus, we attempt to partition the face images merely based on the pairwise similarities; no other assumptions, including the number of identities is used.

\subsection{Problem Formulation}
\label{sec:formulation}

Given a dataset $X$ of size $N$, where each $X_i, i=1,2,...N$ is a data point, we want to directly estimate an $N \times N$ adjacency matrix $Y$, where $Y_{ij}$ is a binary variable indicating whether $X_i$ and $X_j$ are assigned to the same cluster. Assuming that we are given the pairwise conditional probability $p(Y_{ij}|X_i,X_j)$ for all pairs, the goal is to find the overall adjacency matrix $Y$ by maximizing the posterior probability $p(Y|X)$. To model this conditional distribution, we need to consider the dependencies between different variables $Y_{ij}$, for which we propose a triplet interaction constraint to constrain the adjacency matrix $Y$ to be \emph{valid}. By \emph{valid}, we mean that the corresponding graph of that adjacency matrix is transitive and represents a valid partition. This leads to a structured prediction problem, and we use a Conditional Random Field (CRF) model to formulate it and maximize the posterior probability:
\begin{equation}
p(Y|X) = \frac{1}{Z}\prod_{i<j}{\psi_u(Y_{ij})} \prod_{i<j<k}{\psi_t(Y_{ij},Y_{ik},Y_{jk})},
\label{eq:posterior}
\end{equation}
where $Z$ is the normalizing factor, the unary association potential $\psi_u(Y_{ij})=p(Y_{ij}|X_i,X_j)$ is the pairwise conditional distribution over observations and $\psi_t(Y_{ij},Y_{ik},Y_{jk})$ is the triplet interaction potential to constrain $Y$ to be valid. Because the adjacency matrix is symmetric, we only need to take $Y_{ij}$ with $i<j$ as variables, so there are in all $\frac{1}{2}N(N-1)$ output nodes. The unary potential is the likelihood of a pair of data points belonging to the same class, which is exactly what the pairwise similarity stands for, so we apply a transformation to the cosine similarity between the deep representations of two faces to attain the genuine unary potential $\psi_u(Y_{ij}=1)$. In practice, we find that this works well, even without attempting to explicitly model the probability distribution for unary potential.

\begin{figure}[t]
\center
\includegraphics[width=\columnwidth]{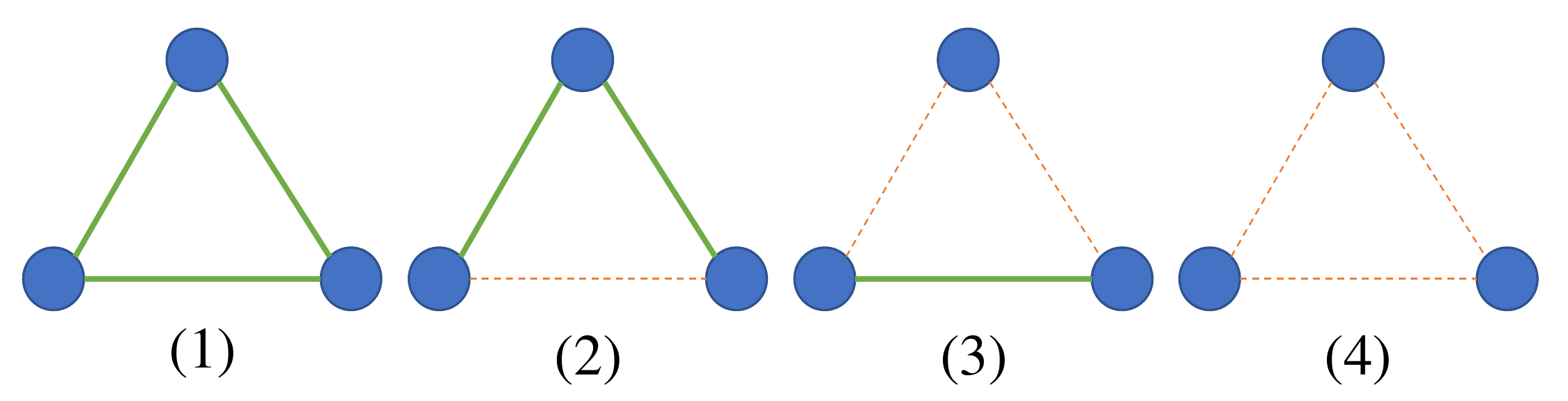}
\caption{Four cases of pairwise adjacency in a triplet. A green line means the two points are connected, i.e. assigned to the same cluster. And a red dash line means the two points are not connected. A valid partition is obtained if and only if none of the triplets are in case (2).}
\label{fig:triplet}
\end{figure}

For a partitional clustering, if a point $i$ is connected to any point $j$ in a cluster, it should also connect to all the other points in that cluster but not to any point outside the cluster. However, not every adjacency matrix satisfies this requirement. In order to check the validity of an adjacency matrix, we use a measure based on triplet consistency. Consider a triplet of any three points, as shown in Figure~\ref{fig:triplet}. Then there are four possible cases regarding the states of three pairs in one triplet.
An adjacency matrix is valid if and only if none of the triplets is in case (2). Hence we can model the dependencies between different $Y_{ij}$ with triplet cliques. In our undirected graph, every triplet $(Y_{ij},Y_{ik},Y_{jk})$ is fully connected, and forms a clique. The interaction potential for a triplet clique is defined as:
\begin{equation}
\psi_t(Y_{ij},Y_{ik},Y_{jk}) = \exp{(-\alpha V(Y_{ij},Y_{ik},Y_{jk}))},
\label{eq:triplet_potential}
\end{equation}
where the energy function $V$ is an indicator function which is $1$ \textit{iff} the triplet is inconsistent and $0$, otherwise:
\begin{equation}
\begin{aligned}
V(Y_{ij},Y_{ik},Y_{jk}) = &(1-Y_{ij})Y_{ik}Y_{jk}+Y_{ij}(1-Y_{ik})Y_{jk}\\
           &+Y_{ij}Y_{ik}(1-Y_{jk})
\end{aligned}
\end{equation}

To seek a valid partition, we consider $\alpha$ in Equation~(\ref{eq:triplet_potential}) to be sufficiently large such that it dominates the formula. However, it is worth noting here that we do not need to explicitly define $\alpha$ in our algorithm, as shown in the next subsection.

\begin{figure*}[t]
\centering
\subfigure[Factor graph]{
    \includegraphics[width=.3\textwidth]{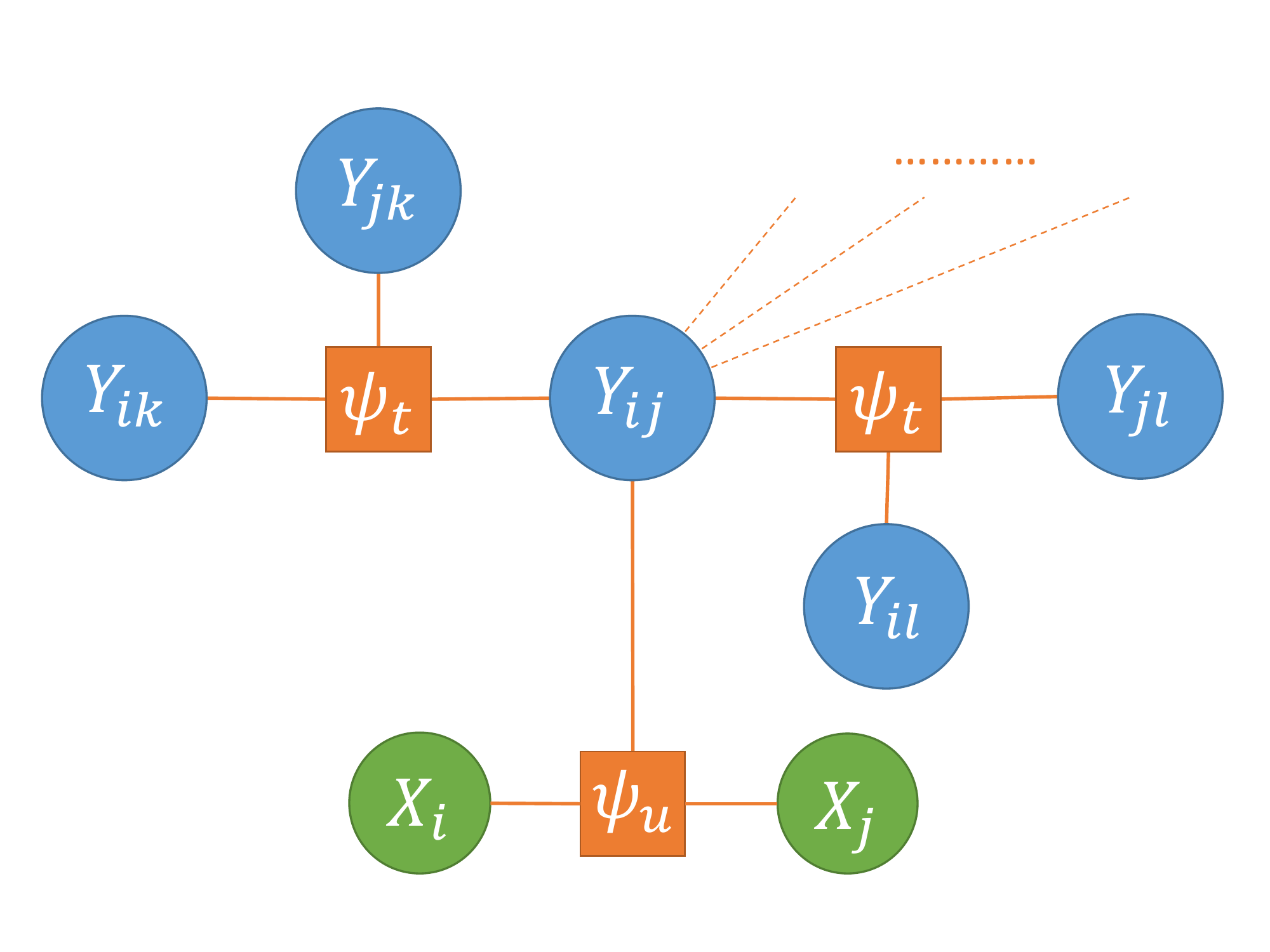}
}
\subfigure[Initialization]{
    \includegraphics[width=.3\textwidth]{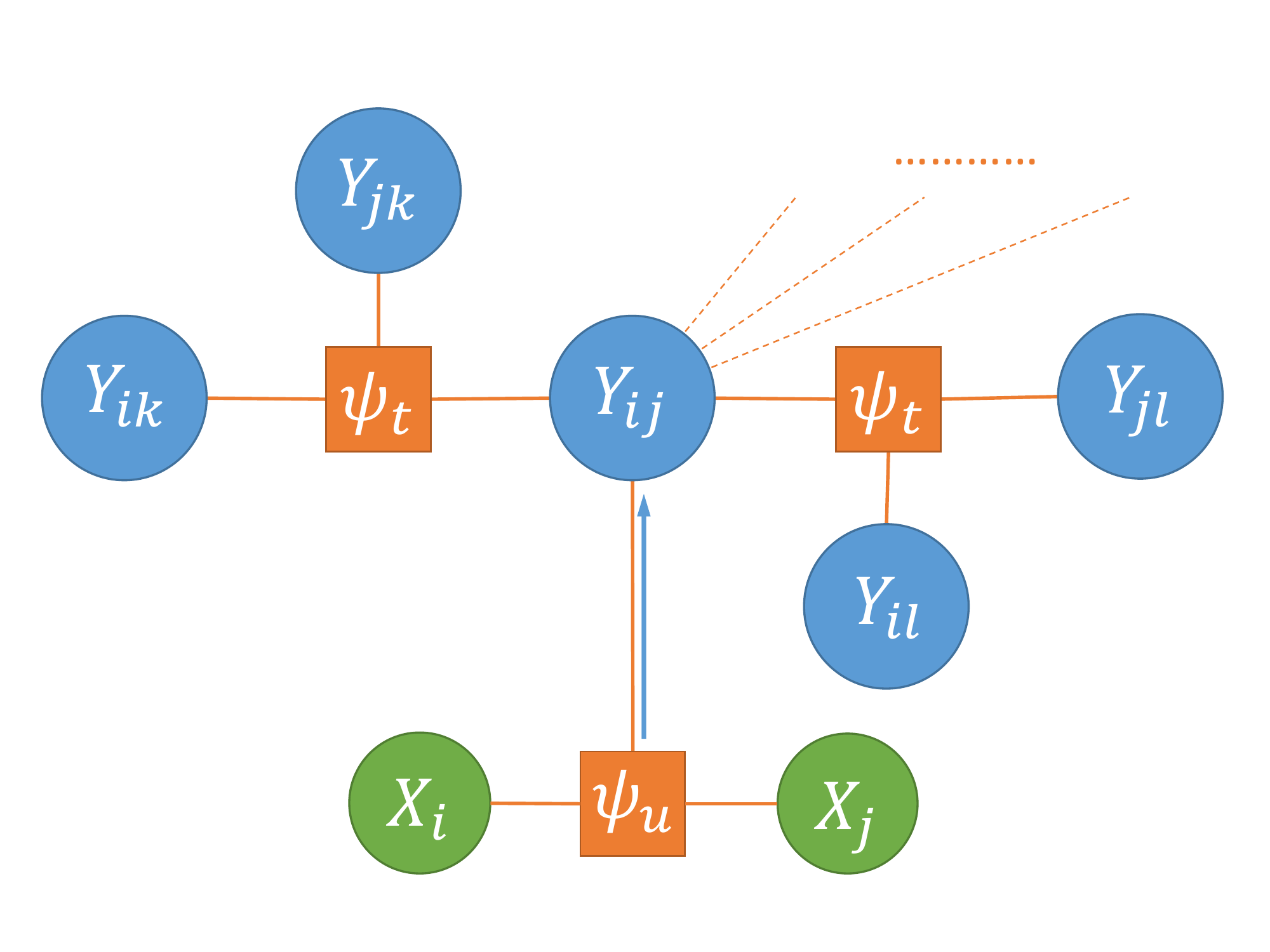}
}
\subfigure[Iteration]{
    \includegraphics[width=.3\textwidth]{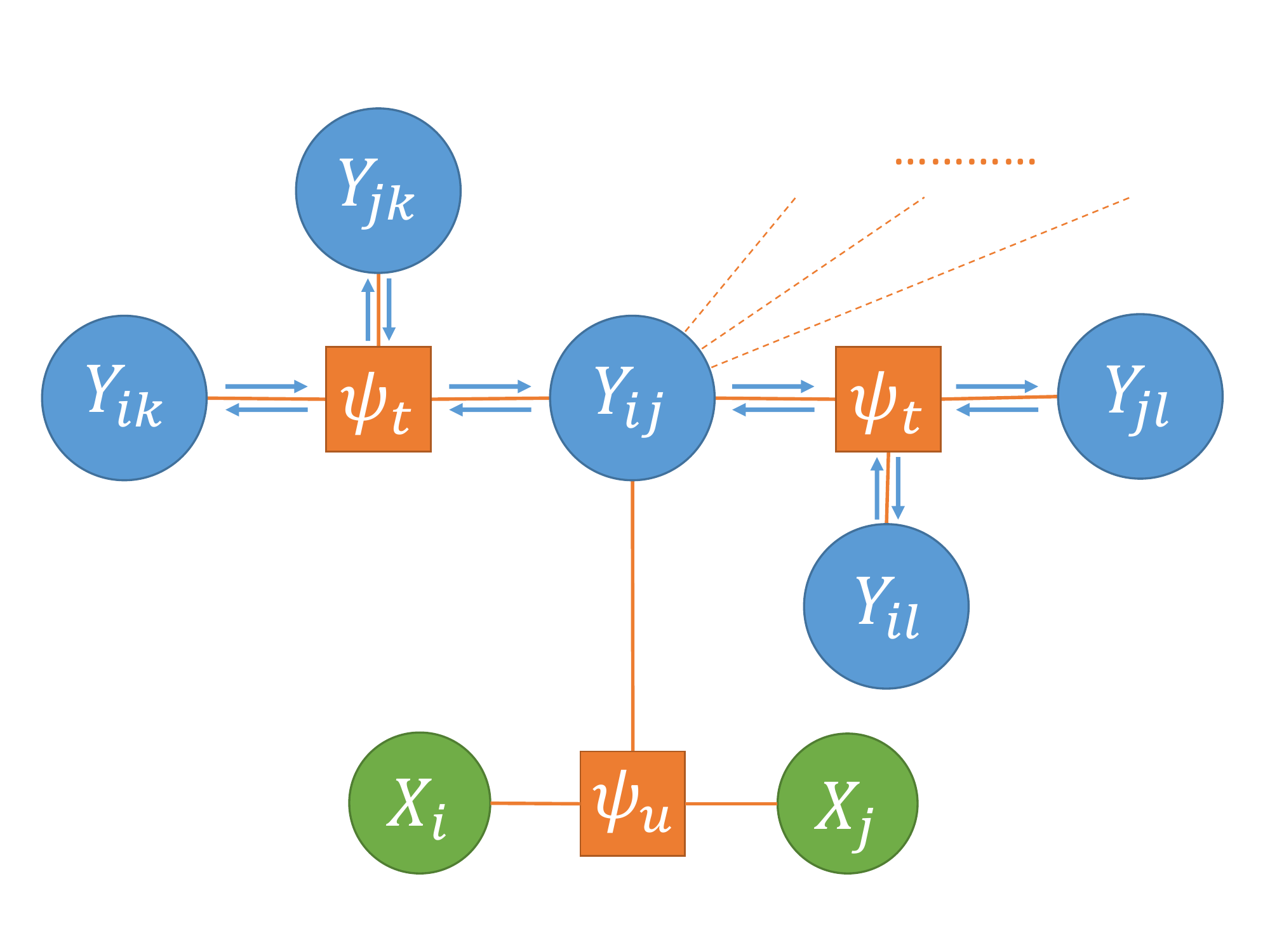}
}
\caption{A graphical illustration of the proposed Conditional Random Field using the neighborhood of output node (pair) $Y_{ij}$ as an example. The figure shows how the nodes are connected, how the factors are related to potentials and how the messages are passed in the graph. Each green node is an input node corresponding to one data point, each blue node is an output node corresponding to an element in the adjacency matrix,  and each rectangle is a factor node representing a potential function, which encodes the constraint between variables. The dash lines represent the omitted links in this figure. There are two kinds of constraints: (i) unary potential which pushes the output to conform with the pairwise similarities and (ii) interaction potential which forces output nodes to be consistent so that $Y$ is valid. During the optimization, messages are propagated among output nodes to directly approach a valid adjacency matrix $Y$ which is mostly consistent with the similarity information.}
\label{fig:graph_examples}
\end{figure*}

Due to numerical issues, usually we take the negative logarithm on both sides of Equation~(\ref{eq:posterior}) and minimize its corresponding energy function:
\begin{equation}
{E(Y,X)} = {\sum_{i<j}{D(Y_{ij})} + \sum_{i<j<k}{\alpha V(Y_{ij},Y_{ik},Y_{jk})}},
\end{equation}
where $D(Y_{ij})=-\log{\psi_u(Y_{ij})}$ is the unary potential energy.

The graph structure of the model is illustrated in Figure~\ref{fig:graph_examples}. Each output node $Y_{ij}$ is in $N-1$ cliques: one association clique with input pair $X_i$ and $X_j$, and $N-2$ interaction cliques consisting of $Y_{ij}$, $Y_{ik}$ and $Y_{jk}$.

\subsection{Inference By Belief Propagation}
\label{sec:lbp}
With the factor graph and potentials defined, we can derive a message formula based on the min-sum algorithm, which is the equivalent for max-product algorithm when working with energy functions~\cite{wiberg1996codes}. We define a message $a_{ij}(Y_{ij})$ as a function of variable $Y_{ij}$, representing the accumulated energy so far for each state of the variable $Y_{ij}$. The main procedure of our algorithm is as follows:

1. Initialize all messages as:
\begin{equation}
a^{0}_{ij}(Y_{ij}) = D(Y_{ij}),
\end{equation}
which is the message sent from the unary potential factor.

2. At iteration $t=1,2,...T$, update the messages as:
\begin{equation}
\begin{aligned}
a^{t}_{ij}(Y_{ij}) = \sum_{k\in N^{t-1}(i,j)} & \min_{Y_{ik},Y_{jk}} (a^{t-1}_{ik}(Y_{ik}) + a^{t-1}_{jk}(Y_{jk}) \\
 & + \alpha V(Y_{ij},Y_{ik},Y_{jk})),
\end{aligned}
\label{eq:message.update}
\end{equation}
where we are summing up the messages from different factors given a state of $Y_{ij}$. Within the sum, we are minimizing over different states of $Y_{ik}$ and $Y_{kj}$.

3. The final state of the variable is determined by:
\begin{equation}
\hat{Y}_{ij}=\argmin_{Y_{ij}}{a^{T}_{ij}(Y_{ij})},
\label{eq:message.decision}
\end{equation}

Here, $N^{t-1}(i,j)$ means the set of the points that are adjacent to either $i$ or $j$ at $(t-1)$ iteration, where by adjacent we mean that it has a lower positive energy than the negative one in that iteration, i.e. $a^{t-1}_{ij}(Y_{ij}=1)>a^{t-1}_{ij}(Y_{ij}=0)$. As mentioned in Section~\ref{sec:crf}, belief propagation is an approximation method for maximizing the posterior probability. It is not guaranteed to find a global optimum. Thus, if there are still inconsistent triplets after the third step, a transitive merge\footnote{Linking all the positive pairs to build clusters. This can be done with linear complexity by building a disjoint-set data structure.} is applied to ensure the clustering result $Y$ is valid. There are several issues worth discussing on this procedure:

First, this is not a standard Loopy Belief Propagation algorithm for CRF in two ways: (1) the unary messages are sent only once and (2) the messages are isotropic, i.e. a message sent from a node $Y_{ij}$ is the sum of all messages it receives. We found that these modifications make the algorithm easier to implement, use less memory, converge faster while have little impact on the quality of the results.

Second, the messages could be normalized by subtracting the same value from both states. Theoretically, it makes no difference to the result, but could avoid numerical underflow and provide stability.

Third, the received messages in Equation~(\ref{eq:message.update}) only include those neighbors $k$ that are adjacent to at least one of $i$ and $j$ in the last iteration, because the messages from other neighbors would have the same value in both states. Thus it makes no difference if we ignore those $k\notin N^{t-1}(i,j)$. For the same reason, we only need to update $Y_{ij}$ whose $N^{t-1}(i,j)$ is not empty.

Fourth, as we assume that $\alpha$ is a very large number, all of the cases where $V(Y_{ij},Y_{ik},Y_{jk})=1$ could be ignored when taking the minimum in equation~(\ref{eq:message.update}). For example, for $Y_{ij}=1$, we do not need to consider the cases where $Y_{ik}=0,Y_{jk}=1$ or $Y_{ik}=1$ and $Y_{jk}=0$. In all the other cases, because $V(Y_{ij},Y_{ik},Y_{jk})=0$, $\alpha$ disappears from the formula.

Given the above optimization, along with our use of adjacency lists and an update list, the complexity of the algorithm is $O(TNM^2)$, where $N$ is the number of data points, $T$ is the number of iterations, and $M$ is the maximum degree\footnote{Number of adjacent points.} of any data point in any iteration. But it should be noted that because $M$ is not a fixed number here, in the worst case the complexity could still become $O(TN^3)$. Besides, we only choose to update the pairs that are in at least one inconsistent triplet, so the update list is typically much shorter after several iterations. On the LFW dataset, only $0.59\%$ of the total pairs\footnote{There are in all $87,549,528$ pairs in the LFW dataset.} are updated, and no more than $1,000$ pairs are in the update list after the fourth iteration.

\begin{figure}[t]
\centering
\subfigure[noisy circles]{
    \includegraphics[width=.29\linewidth]{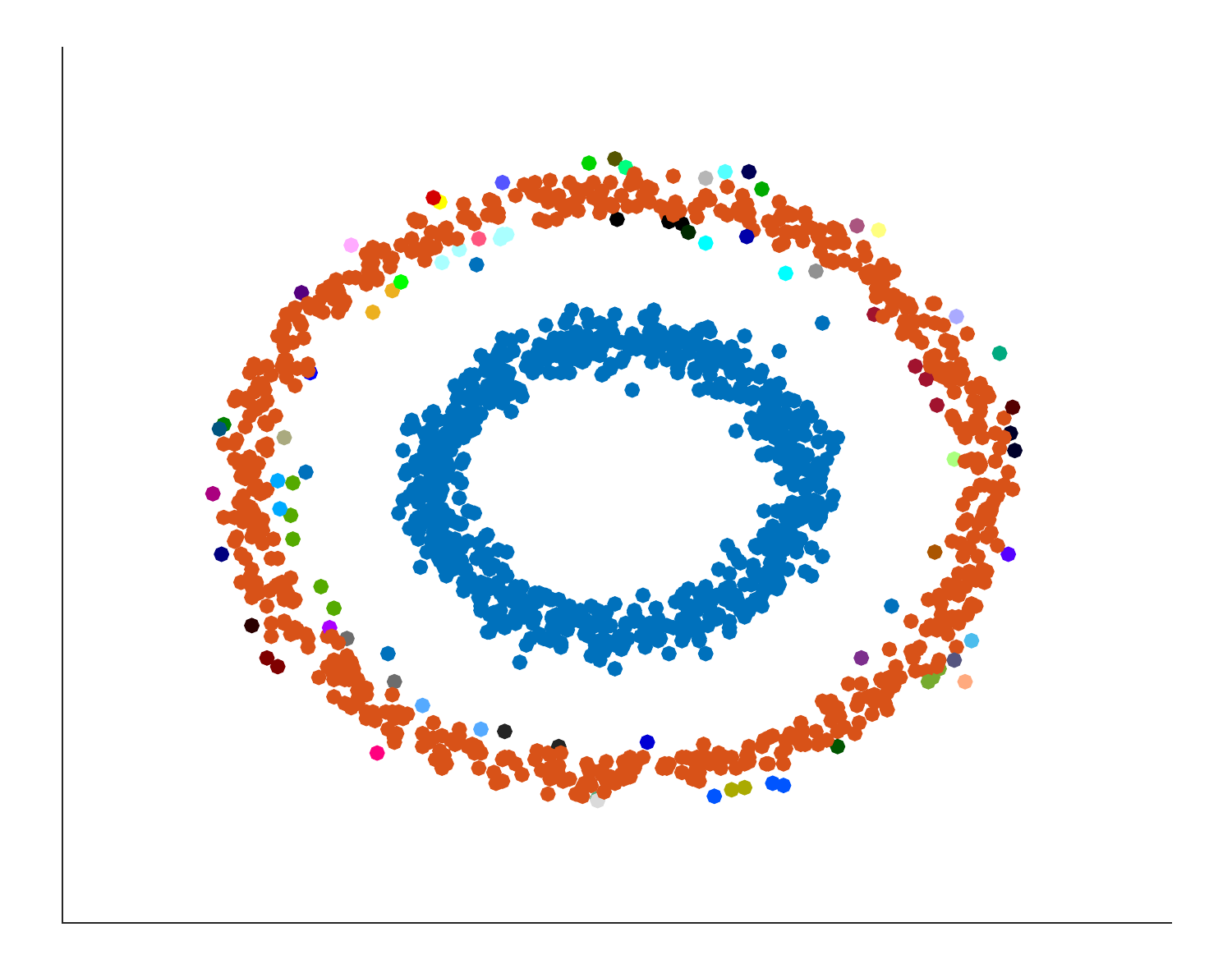}
}
\subfigure[noisy moons]{
    \includegraphics[width=.29\linewidth]{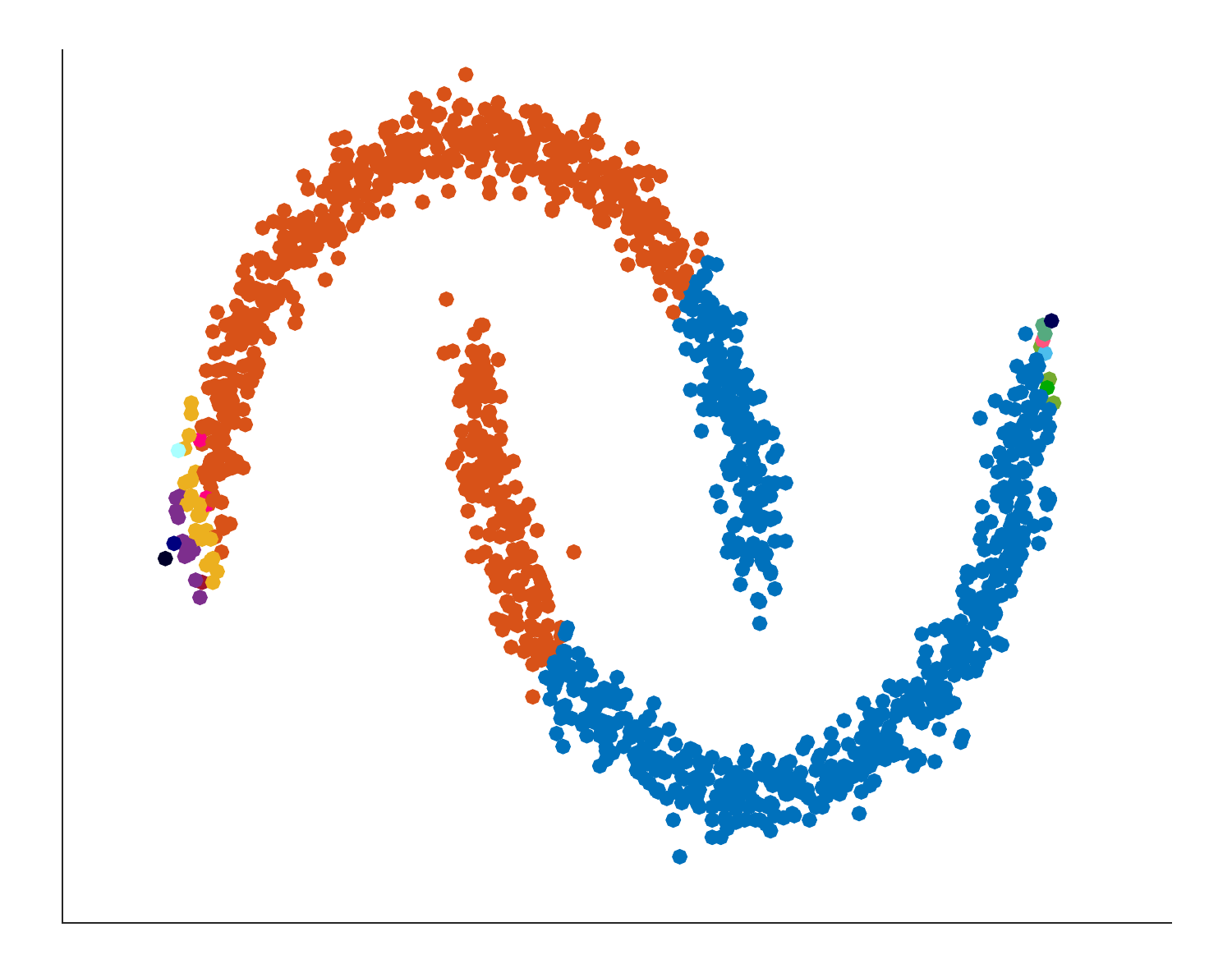}
}
\subfigure[varied]{
    \includegraphics[width=.29\linewidth]{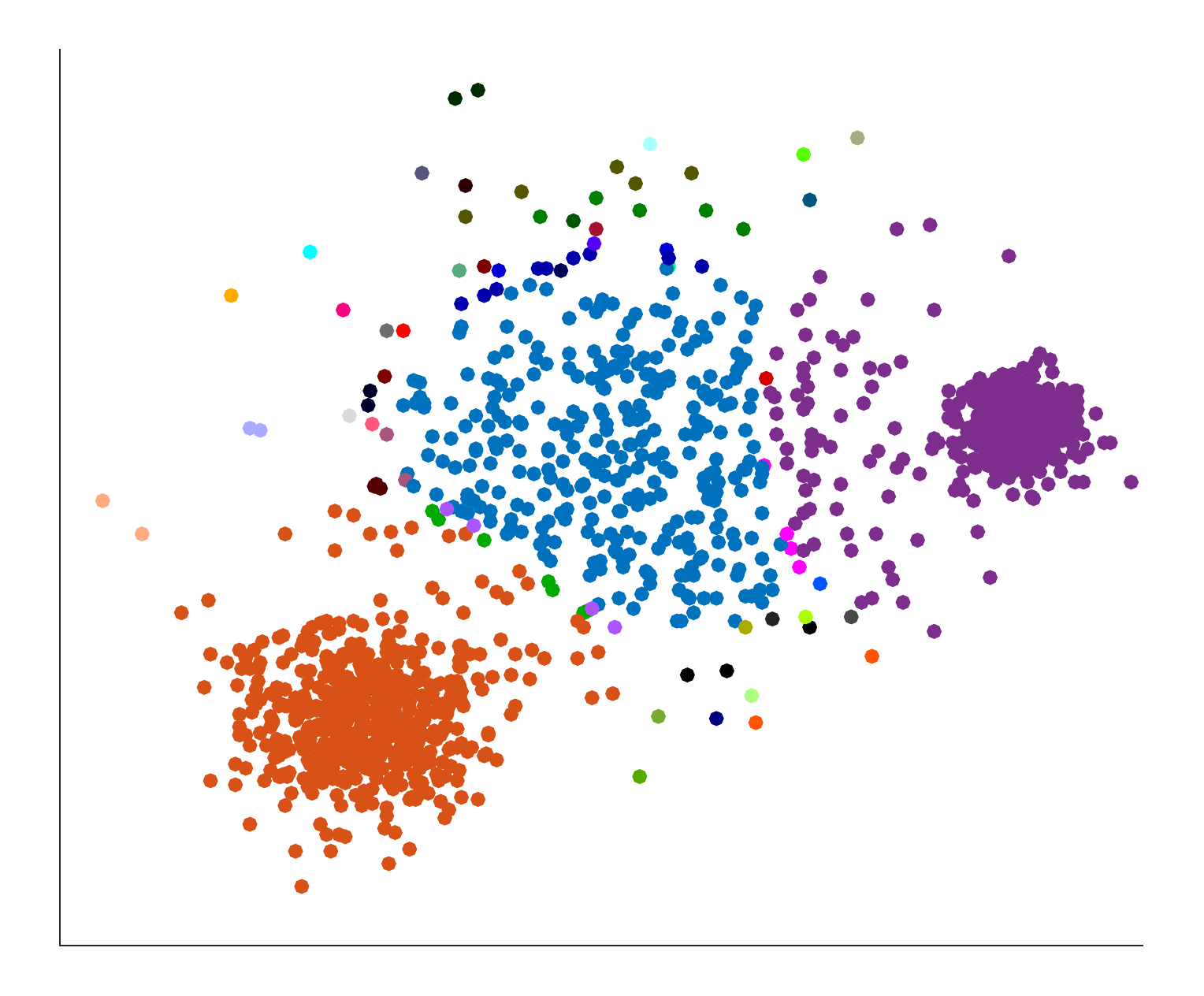}
}\\
\subfigure[aniso]{
    \includegraphics[width=.29\linewidth]{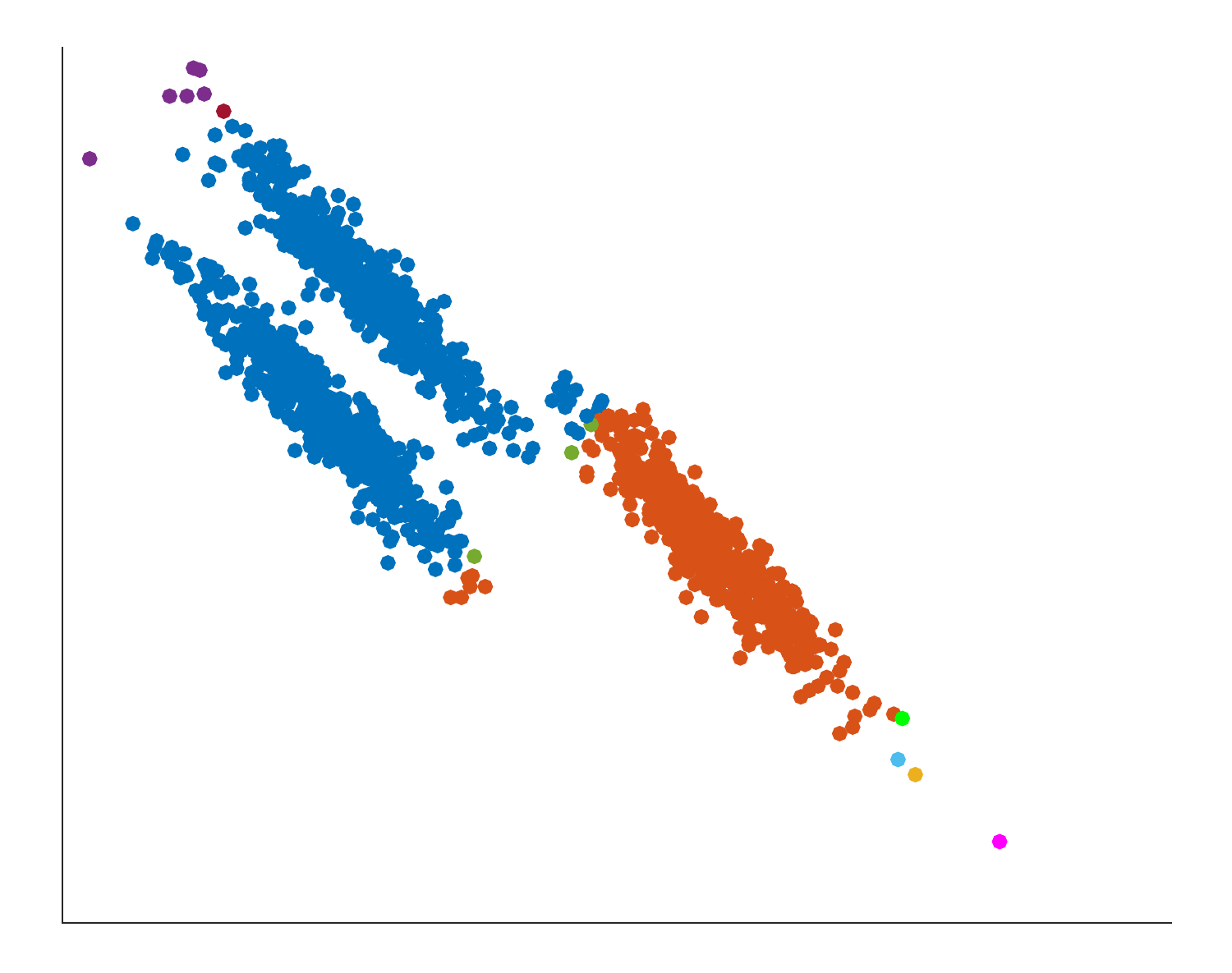}
}
\subfigure[blobs]{
    \includegraphics[width=.29\linewidth]{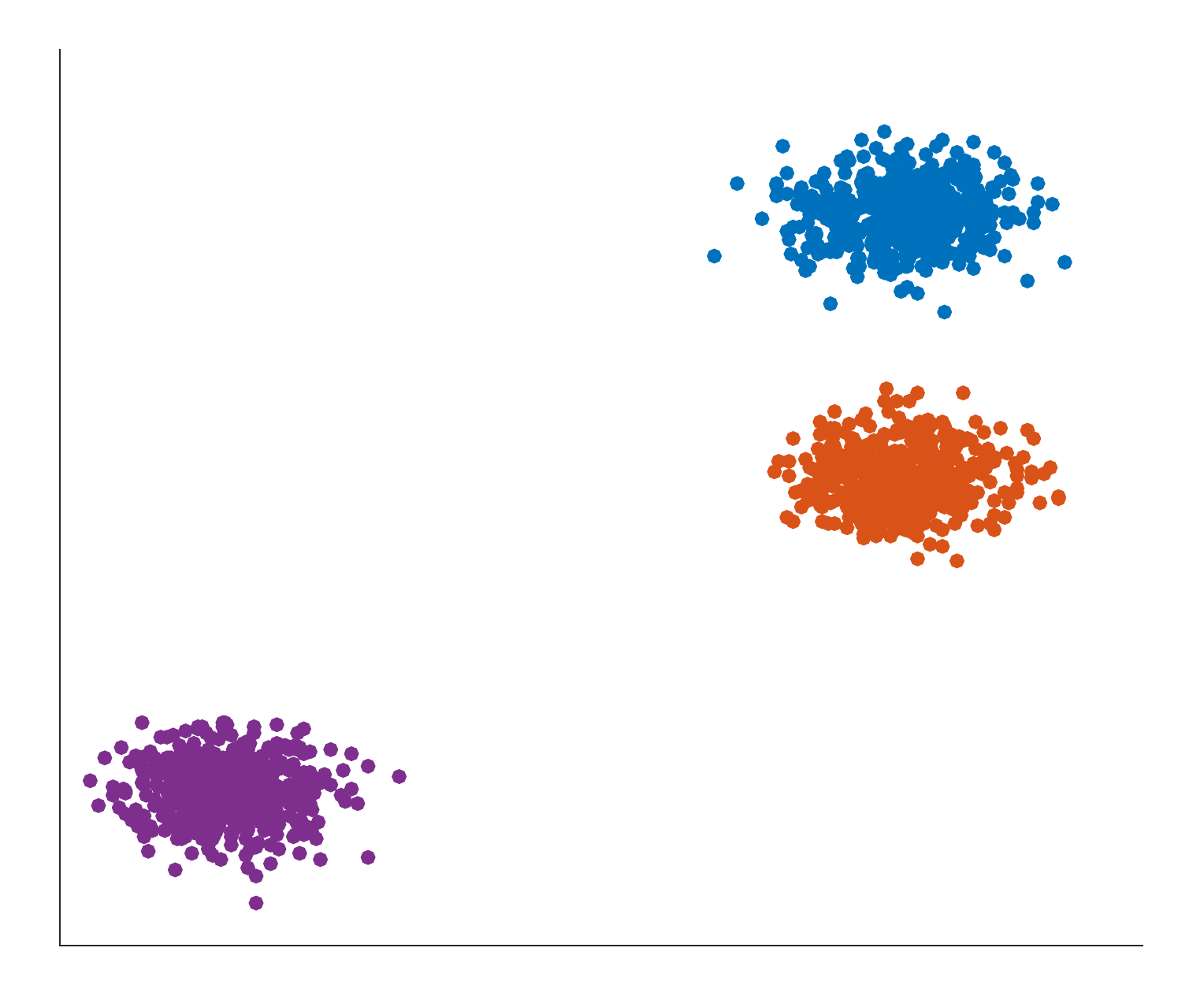}
}
\subfigure[no structure]{
    \includegraphics[width=.29\linewidth]{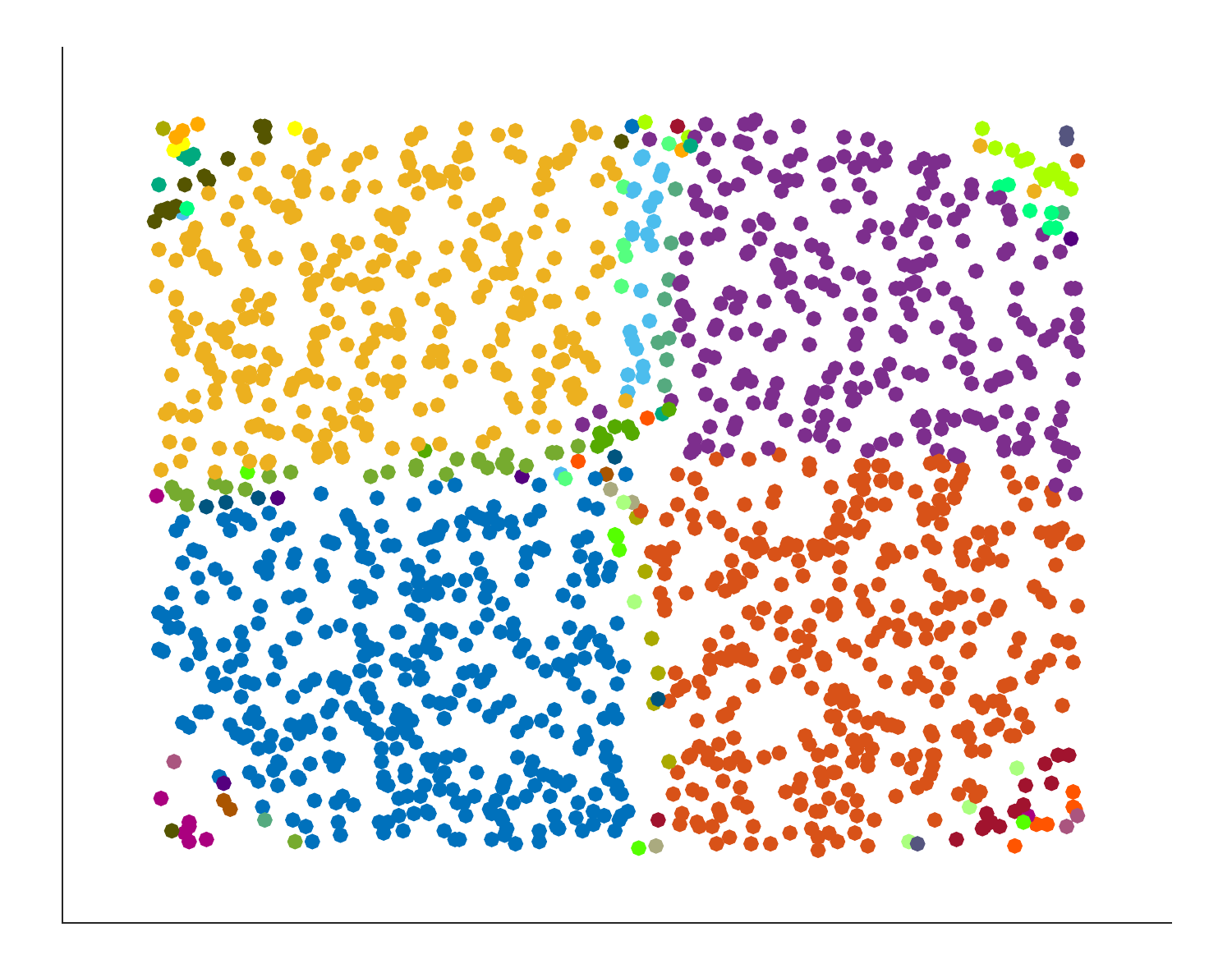}
}\\

\caption{The result of proposed clustering on toy examples from scikit-learn using a Radial Basis Function (RBF) kernel. We tune the parameter $\gamma$ according to the datasets. The colors indicate the assigned label for each node.}
\label{fig:toy}
\end{figure}

We test the proposed clustering algorithm on a set of toy examples from the scikit-learn library\footnote{\url{http://scikit-learn.org/stable/auto_examples/cluster/plot_cluster_comparison.html}}. The results are shown in Figure~\ref{fig:toy}. Because we use the Radial Basis Function (RBF) kernel for the similarity metric on the toy examples, the data points are mainly grouped according to the Euclidean distances between them. Although the number of clusters is not specified in our algorithm, the main groups can be recovered in the belief propogation procedure. The outliers are often assigned as small, separate clusters.

\begin{figure}[t]
\center
\includegraphics[width=0.85\columnwidth]{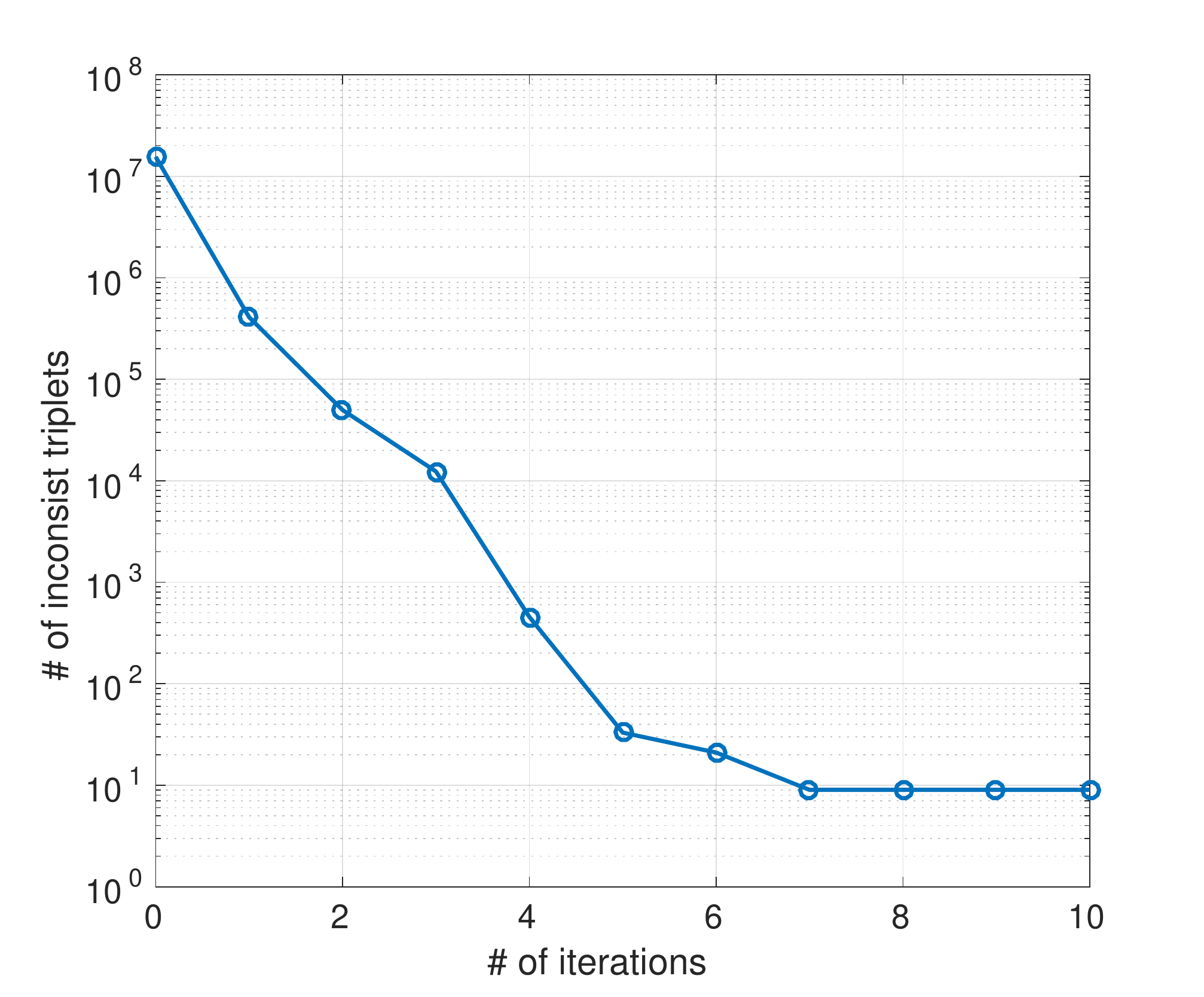}
   \caption{Number of inconsistent triplets on LFW at each iteration. It decreases rapidly until convergence.}
\label{fig:loss_lfw}
\end{figure}

Figure~\ref{fig:loss_lfw} shows how the number of inconsistent triplets decreases with iterations during clustering of the $13,233$ images in the LFW dataset. The model converges rather quickly to a stage where only a small number of inconsistent triplets remain. Because the number of remaining inconsistent triplets is usually very small, we do not explicitly force convergence to $0$ but apply a transitive merge on the current adjacency matrix to attain the final valid clustering. On LFW, only 6 pairs change their states after we apply transitive merge.

\subsection{Semi-supervised clustering}
\label{sec:semi}

In semi-supervised or constrained clustering, we utilize the given side information, usually in the form of ``must-link" pairs and ``cannot-link" pairs. These pairs can either be specified by users or automatically generated with another algorithm to improve clustering performance. The must-links specify pairs of face images that belong to the same identity, while the cannot-links specify pairs that belong to different identities. One way to make use of these pairs is to propagate the constraints. Because our framework is optimized by propagating messages, it becomes quite straight forward to incorporate these constraints: we change the unary potentials of the constrained pairs based on the side information provided. If we are very confident with the given constraints (as is the case in our experiments which use ground-truth labels for side information), we can set the positive unary potentials as $1$ for must-link and $0$ for cannot-link constraints, resulting in very large unary energies. Equation (\ref{eq:message.update}) states that very high energy would be avoided when passing messages so the model can still be optimized under these constraints.

\subsection{Efficient Variant of the clustering algorithm using \textit{k}-NN Graph}

The complexity of the proposed clustering algorithm depends on the degrees of data points during the optimization. However, since this number is not fixed, in the worst case its complexity could still be close to $O(TN^3)$, where $T$ and $N$ are the number of iterations and data points, respectively. Therefore, we propose a variant of the algorithm which has a fixed linear complexity. The idea is similar to the one in~\cite{otto2017clustering}, which takes advantage of approximate \textit{k}-Nearest Neighbors (\textit{k}-NN) methods. Instead of estimating the adjacency of every pair within the dataset, we optimize the joint posterior probability of all $Y_{ij}$ where $(i,j)$ is an edge in the \textit{k}-NN graph. These $Y_{ij}$ compose a subset of elements in the full adjacency matrix $Y$. The same procedure outlined in section~\ref{sec:lbp} can still be used for optimization with a few modifications: (1) The neighbor list $N(i)$ now is a fixed list given by the approximate \textit{k}-NN method, (2) we only update $Y_{ij}$ where $i\in N(j)$ or $j\in N(i)$, and (3) we only need to compute the unary potentials in the initialization step for pairs which will be used in the next iteration, i.e. they are neighbors or they have at least one shared neighbor. 
While time complexity of this variant, given a pre-computed \textit{k}-NN graph, is also $O(TNM^2)$ as in section~\ref{sec:lbp}, $M$ is now a fixed number. In particular, we use the same approximate \textit{k}-NN method, \textit{k}-d tree, with the same configuration as in~\cite{otto2017clustering} to build the \textit{k}-NN graph. The complexity of building \textit{k}-NN graph is $O(N\log{N})$ with a fixed search size. If we increase the search size linearly with the size of the dataset, as in~\cite{otto2017clustering}, it leads to $O(N^2)$ complexity. We use a fixed search size of $2,000$ in our experiments.

\section{Experimental Results}

\subsection{Face Representation Performance}

\newcounter{magicrownumbers}
\newcommand\rownumber{\stepcounter{magicrownumbers}(\arabic{magicrownumbers})}

\begin{table}[t]
\centering

\caption{BLUFR verification performance on LFW. ResNet was trained on the combined VGG+CASIA-Webface dataset. Verification Rates (VR) at a False Alarm Rate (FAR) are reported as (mean - standard deviation) across 10 folds.}
\begin{tabularx}{\columnwidth}{Xr}
\toprule
Network & VR@FAR=$0.1\%$ \\
\midrule
50-Layer Pre-activated ResNet &  $91.04$\%\\
50-Layer Pre-activated ResNet, 10-crop &  $92.22$\% \\
101-Layer Pre-activated ResNet &  $91.18$\%\\
101-Layer Pre-activated ResNet, 10-crop &  $92.10$\% \\ 
\bottomrule

\end{tabularx}
\label{tab:blufr_ver}
\end{table}

We evaluate the performance of our representation using ResNet on the BLUFR protocol~\cite{BLUFR}. Because the performance of state-of-the-art face representations on standar LFW verification protocol has saturate, Liao et al.~\cite{BLUFR} made use of the entire LFW dataset to design the BLUFR protocol. In this protocol, a 10-fold cross-validation test is defined for both \emph{face verification} and \emph{open-set face identification}. For \emph{face verification}, a Verification Rate (VR) is reported for each split with a false alarm rate: FAR$=0.1\%$. For \emph{open-set identification}, Detection and Identification Rate (DIR) at Rank-1 corresponding to FAR$=1\%$ is computed.

Table~\ref{tab:blufr_ver} gives a summary of our face representation performance on the BLUFR protocol's verification experiment~\cite{BLUFR}. We trained $50$ and $101$-layer fully pre-activated ResNets on the combined VGG and CASIA-Webface datasets (using our cleaned version of VGG). A subset o $1,000$ images are randomly selected from CASIA-Webface are kept as the validation set. The $50$-layer network achieves a $91.04\%$ verification rate at $0.1\%$ FAR after training for $37$ epochs, at which point the classification accuracy on the validation set stabilizes. Increasing the network depth to 101 layers does not result in a performance improvement. Using the 10-crop strategy leads to a minor performance improvement (approximately 1\% VR at 0.1\% FAR), at the cost of substantially increased feature extraction time.

Our best results on  BLUFR in Table~\ref{tab:blufr_ver} are: $92.22\%$ VR at $0.1\%$ FAR for verification, and $62.05\%$ DIR at $1\%$ FAR for open-set identification. This is comparable to some newly reported results on the protocol. For example, Cheng at al.~\cite{cheng2016bootstrapping} used a GoogLeNet-style Inception architecture combined with traditional Joint-Bayes and attained a $92.19\%$ VR at $0.1\%$ FAR. They further improved this result to $93.05\%$ using their method for estimating the Joint-Bayes parameters. Lv et al.~\cite{lv2016landmark} proposed a data augmentation method (perturbing detected facial landmark locations, prior to image alignment), again using an Inception architecture, and attained a $63.73\%$ DIR at 1\% FAR in open-set identification, using the fusion of 3 models (the best single-model performance is $57.90\%$). These results indicate that our results could potentially be further improved, through the incorporation of metric learning methods, or fusing multiple models.

\subsection{Face Clustering}

\begin{figure}[t]
\center
\includegraphics[width=0.7\columnwidth]{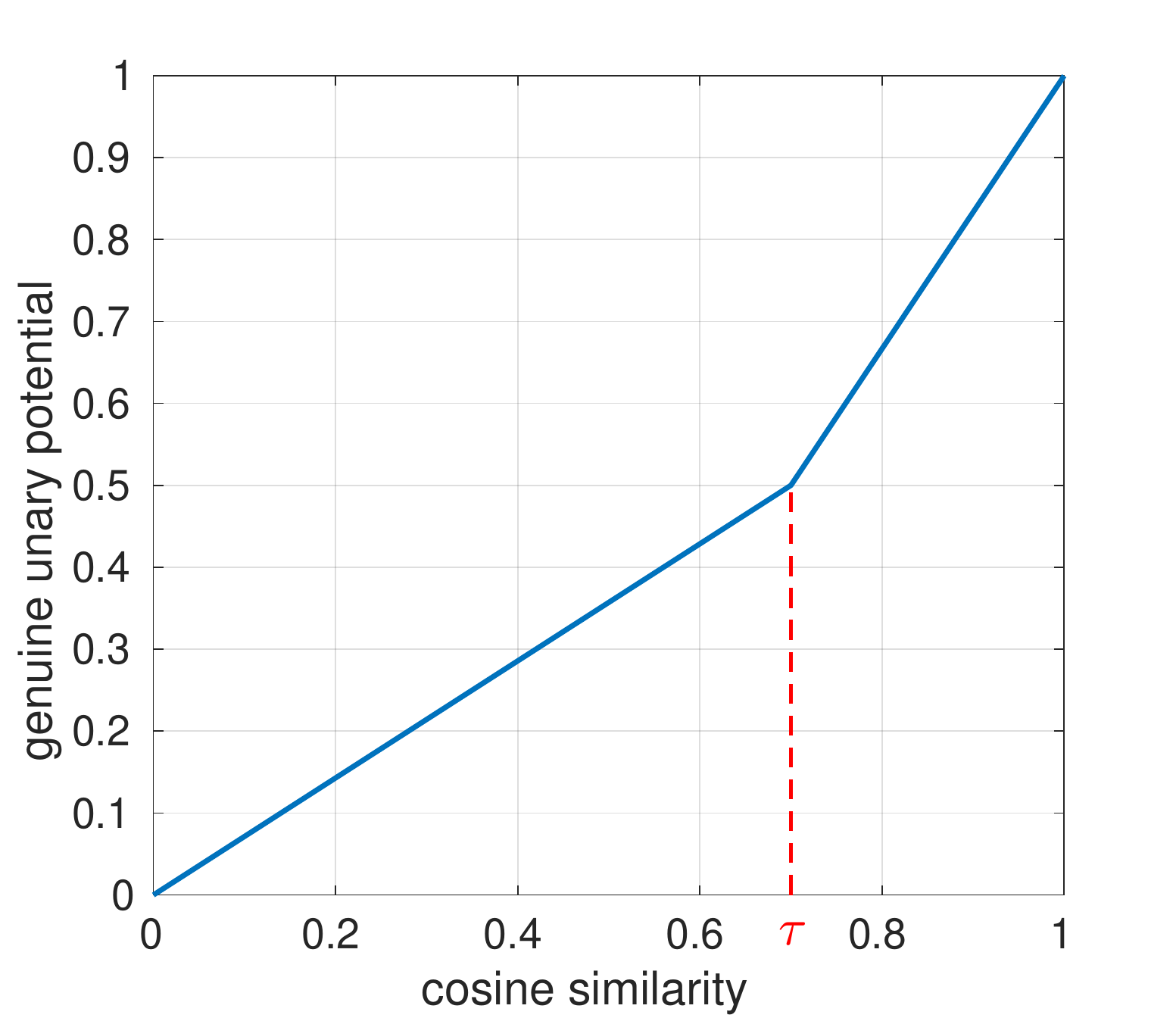}
\caption{Transformation function used to map the cosine similarity to the genuine unary probability $\psi_u(Y_{ij}=1)$. A threshold $\tau$ is used to split the function into two pieces. The cosine similarity is non-negative because the the features are extracted from the ReLU layer of the last hidden layer in the ResNet.}
\label{fig:transform_cosine}
\end{figure}

We use the $50$-layer ResNet architecture, with $10$-crop (Table~\ref{tab:blufr_ver}) strategy as the representation for our clustering experiments. We evaluate our clustering algorithm on two unconstrained face datasets (LFW and IJB-B). Before applying the message passing procedure, we obtain unary potentials described in~\ref{sec:formulation} using a transformation function shown in Figure~\ref{fig:transform_cosine}. Threshold $\tau$ is the only parameter throughout our experiments. Different $\tau$ values control the balance between the recall rate and the precision rate of the resulting partition, which are defined in~\ref{sec:measures}. But it is worth emphasizing that the transformation function itself (Figure~\ref{fig:transform_cosine}) is not a necessary part of the clustering algorithm and one may choose other ways to initialize the unary potential. We use this transformation function because it is easy to compute and work well empirically. The same threshold $\tau=0.7$ is used as default unless otherwise stated.

We call our algorithm Conditional Pairwise Clustering (ConPaC), which is implemented in C++ and evaluated on an Intel Xeon CPU clocked at 2.90GHz using 24 cores. The ConPac algorithm is compared to the following benchmarks: (1) \textit{k}-means, (2) Spectral Clustering~\cite{ng2001spectral}, (3) Sparse Subspace Clustering (SSC)~\cite{elhamifar2009sparse} (4) Affinity Propagation~\cite{frey2007clustering}, (5) Agglomerative Clustering, (6) Rank-order clustering~\cite{zhu2011rank}, and (7) Approx. Rank-order clustering~\cite{otto2017clustering}. We use the MATLAB R2016a implementation of the \textit{k}-means algorithm, and a third-party MATLAB implementation of Spectral Clustering\footnote{\url{http://www.mathworks.com/matlabcentral/fileexchange/34412-fast-and-efficient-spectral-clustering/content/files/SpectralClustering.m}}. We use the author's implementation of SSC~\footnote{\url{http://www.vision.jhu.edu/code/}}. For Affinity Propagation and Agglomerative Clustering, we use the scikit-learn implementation~\cite{scikit-learn}. Since all the implementations tested use built-in functions for the core steps, such as matrix decomposition, the difference in the programming language shouldn't affect the run-time excessively, and we consider it fair to compare the execution times of these implementations.

We use the same representation for all the algorithms except Approx. Rank-order, for which we follow~\cite{otto2017clustering} because it generates better clustering results. We use Euclidean distance for \textit{k}-means, RBF kernel for Affnity Propagation, and cosine similarity for Spectral Clustering.

\begin{figure*}[t]
\center
\includegraphics[width=1.0\linewidth]{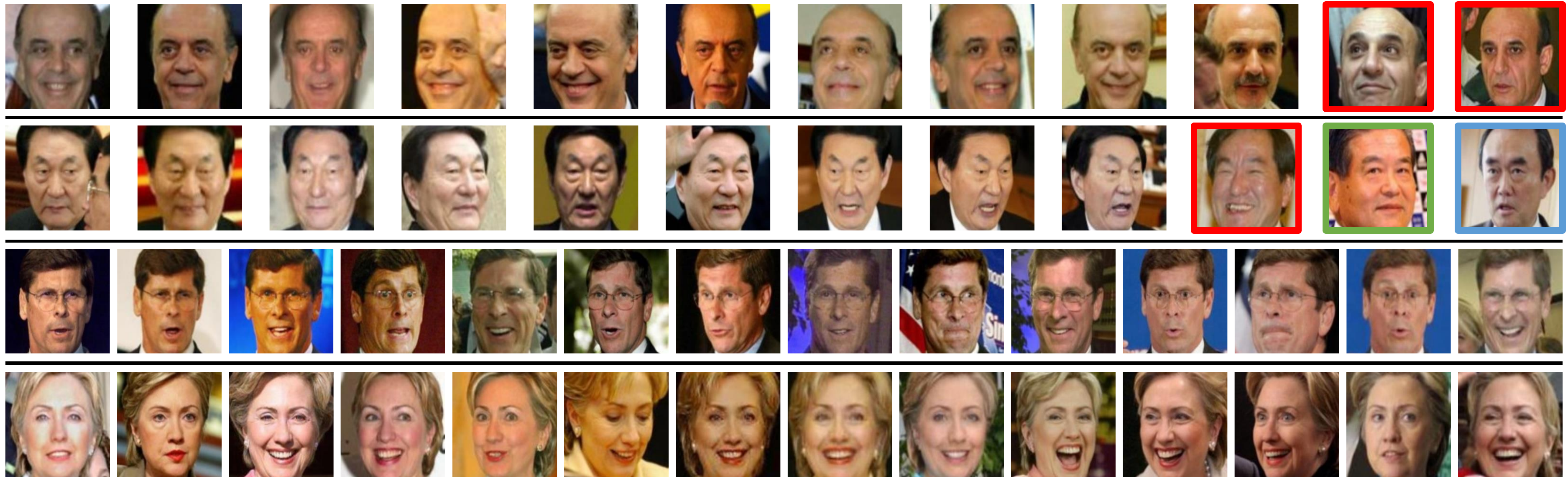}
   \caption{Example clusters by the proposed clustering algorithm on the LFW dataset. The first two rows show example impure clusters while the other two rows show pure clusters. For impure clusters, different colors of bounding boxes indicate the images are from different identities, and images without bounding boxes are from the same identity.}
\label{fig:example_lfw}
\end{figure*}

\subsubsection{Evaluation Measures}
\label{sec:measures}
Two measures are used to evaluate the clustering results, \textit{Pairwise F-measure} and \textit{BCubed F-measure}. Both compute a \textit{F-score}, which is the harmonic mean of \textit{Precision} and \textit{Recall}. The difference between them lies in the metrics used for precision and recall. 

In \textit{Pairwise F-measure}, \textit{Precision} is defined as the fraction of pairs that are correctly clustered together over the total number of pairs that belong to the same class. \textit{Recall} is defined as the fraction of pairs that are correctly clustered together over the total number of pairs that are in the same cluster. In other words, we are using the labels for all the $\frac{1}{2}N(N-1)$ pairs in the dataset. Thus, we can define the True Positive Pairs ($TP$), False Positive Pairs ($FP$) and False Negative Pairs ($FN$). Then \textit{Precision} and \textit{Recall} can be calculated as:
\begin{equation}
Pairwise\ Precision = \frac{TP}{TP+FP}
\label{eq:pairwise_precision}
\end{equation}
\begin{equation}
Pairwise\ Recall = \frac{TP}{TP+FN}
\label{eq:pairwise_recall}
\end{equation}

\textit{BCubed F-measure}~\cite{amigo2009comparison} defines \textit{Precision} as point precision, namely how many points in the same cluster belong to its class. Similarly, point recall represents how many points from its class appear in its cluster. Formally, we use $L(i)$ and $C(i)$ to, respectively, denote the class and cluster of a point $i$, and define the \textit{Correctness} between two points $i$ and $j$ as:
\begin{equation}
Correctness(i,j) =     
\begin{cases}
  1, & \text{if $L(i)=L(j)$ and $C(i)=C(j)$} \\
  0, & \text{if otherwise}
\end{cases}
\end{equation}
The \textit{Precision} and \textit{Recall} are defined as:
\begin{equation}
BCubed\ Precision = \frac{1}{N}\sum_{i=1}^{N}{\sum_{j\in C(i)}\frac{Correctness(i,j)}{|C(i)|}}
\end{equation}
\begin{equation}
BCubed\ Recall = \frac{1}{N}\sum_{i=1}^{N}{\sum_{j\in L(i)}\frac{Correctness(i,j)}{|L(i)|}}
\end{equation}
where $|C(i)|$ and $|L(i)|$ denote the sizes of the sets $C(i)$ and $L(i)$, respectively.

The F-measure, or F-score for both criteria is given by:
\begin{equation}
F = \frac{2\times Precision \times Recall}{Precision+Recall}
\end{equation}

\textit{Pairwise F-measure} is a more commonly used measure, and \textit{BCubed F-measure} is the formal evaluation measure for the IJB-B dataset. The difference between the two is that \textit{Pairwise F-measure} puts relatively more emphasis on large clusters because the number of pairs grows quadratically with cluster size, while under \textit{BCubed F-measure} clusters are weighted linearly based on their size.

\subsubsection{Evaluation on the LFW Dataset}

LFW is quite an imbalanced dataset, with only 1,680 classes (individuals) containing more than one face. Since we cannot assume that our datasets will be well balanced, we do the experiments on the whole LFW dataset.

\begin{table}[t]
\centering
\caption{Comparison of the F-measures of the proposed algorithm and other clustering algorithms on LFW dataset. The number of identities (true number of clusters) in LFW is $5,749$. Run-time is reported in the format of hh:mm:ss.}
\scriptsize
\begin{threeparttable}
\begin{tabularx}{\columnwidth}{Xrrrr}
\toprule
            & Pairwise  & BCubed    &  \# of    &           \\
Algorithm   & F-measure & F-measure &  Clusters & Run-time  \\
\midrule
TPE~\cite{sankaranarayanan2016triplet}\tnote{$_*$}    & $0.955$   & $-$   & $5,351$   & $-$\\
\textit{k}-means     & $0.098$   & $0.680$   & $5,749$   & $00:04:26$\\
\textit{k}-means     & $0.346$   & $0.468$   & $500$     & $00:00:14$\\
Spectral    & $0.033$   & $0.559$   & $5,749$   & $06:52:22$\\
Spectral    & $0.257$   & $0.249$   & $75$      & $01:00:56$\\
SSC         & $0.186$   & $0.430$   & $500$     & $00:31:52$\\
Affinity
Propagation & $0.320$   & $0.577$   & $1,203$   & $00:06:56$\\
Agglomerative & $0.962$   & $0.892$   & $4,500$       & $00:03:04$\\
Rank-Order  & $0.813$   & $0.891$   & $5,699$   & $00:00:33$\\
Approx.
Rank-Order  & $0.861$   & $0.875$   & $6,801$   & $00:00:12$\\
\midrule
ConPaC 
(proposed)  & $\mathbf{0.965}$   & $\mathbf{0.922}$   & $6,352$   & $00:00:39$\\
\bottomrule
\end{tabularx}
\begin{tablenotes}
\item[$_*$] This result is reported in~\cite{sankaranarayanan2016triplet}.
\end{tablenotes}
\end{threeparttable}
\label{tab:lfw_cluster}
\end{table}

\begin{figure*}[t]
\center
\includegraphics[width=1.0\linewidth]{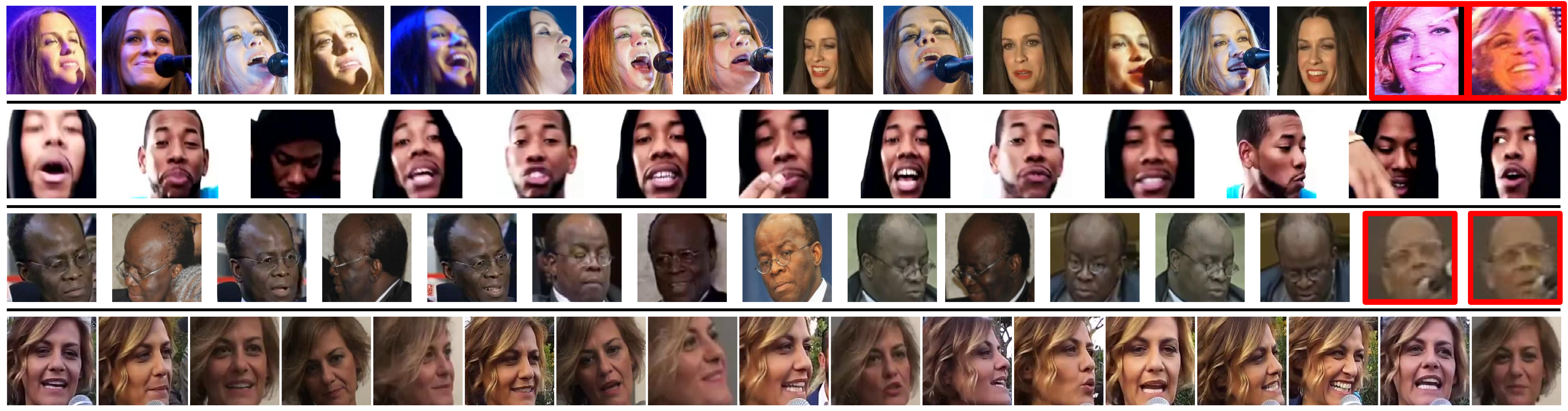}
   \caption{Example clusters by the proposed clustering algorithm on two of the experiments in IJB-B clustering protocol, IJB-B-32 and IJB-B-1024. The first and the second row show an impure and pure cluster on IJB-B-32 dataset, respectively. The third row and the fourth row show an impure and pure cluster on IJB-B-1024 dataset, respectively. For impure clusters, red bounding boxes indicate the images are from a different identity.}
\label{fig:example_cs3}
\end{figure*}

The number of clusters $C$ is dynamically selected during the update of the ConPaC, but it is required as an input parameter for \textit{k}-means and spectral clustering. So we first evaluate their performance with the ground-truth or the true number of clusters $C$, $C=5,749$. Then we repeat the clustering with several different values and report the one that gives the best performance.

Table~\ref{tab:lfw_cluster} shows that the performance of \textit{k}-means and spectral clustering is poor with the ground-truth $C$. This is because these two algorithms do not perform well with unbalanced data. After tuning the parameter $C$, the proposed algorithm still performs better than competing algorithms. Notice that agglomerative clustering using average linkage performs much better than k-means and spectral clustering. This is consistent with our observation in Section IV-B since agglomerative clustering also doesn't assume a balanced dataset and focuses on the pairwise similarities. We also compare the results with that reported in Triplet Probabilistic Embedding (TPE)~\cite{sankaranarayanan2016triplet}, which uses agglomerative clustering on a different representation. For the run-time, since the sizes of clusters in the LFW dataset are mostly very small, the time complexity of ConPaC is low and it take less than one minute to finish. 

Some example clusters by ConPaC for LFW are shown in Figure~\ref{fig:example_lfw}, where the first two rows show two impure clusters while the other two show two pure clusters. Face images in these clusters have different illumination conditions, backgrounds, and poses. Given these challenges, the algorithm still groups most images successfully according to true identities. For the two impure clusters, the mis-grouped images are very similar to other images in the same cluster.

\subsubsection{Evaluation on the IJB-B Dataset}

The results of the 7 experiments in the IJB-B clustering protocol are shown in Table~\ref{tab:cs3_cluster}. Since B-Cubed F-measure is the adopted evaluation measure in the IJB-B protocol, we only report the result with the best B-Cubed F-measure for each algorithm. Because of the memory limit, some clustering algorithms cannot be tested on larger datasets. In such cases, we report the result as ``-". We set $\tau$ as $0.55$, $0.6$ and $0.65$, respectively, for IJB-B-32, IJB-B-64 and IJB-B-128 and $0.7$ for the other datasets.

As the number of identities increases, the F-scores of both competing and the proposed algorithm decrease. While the proposed algorithm shows a significant advantage in terms of F-score on the first few experiments, the gain diminishes as the number of clusters increases. As explained in Section~\ref{sec:quality_sim}, this decrease in performance with an increase in the number of clusters is mainly due to the brittleness of the representation.

\begin{figure}[t]
\includegraphics[width=0.49\columnwidth]{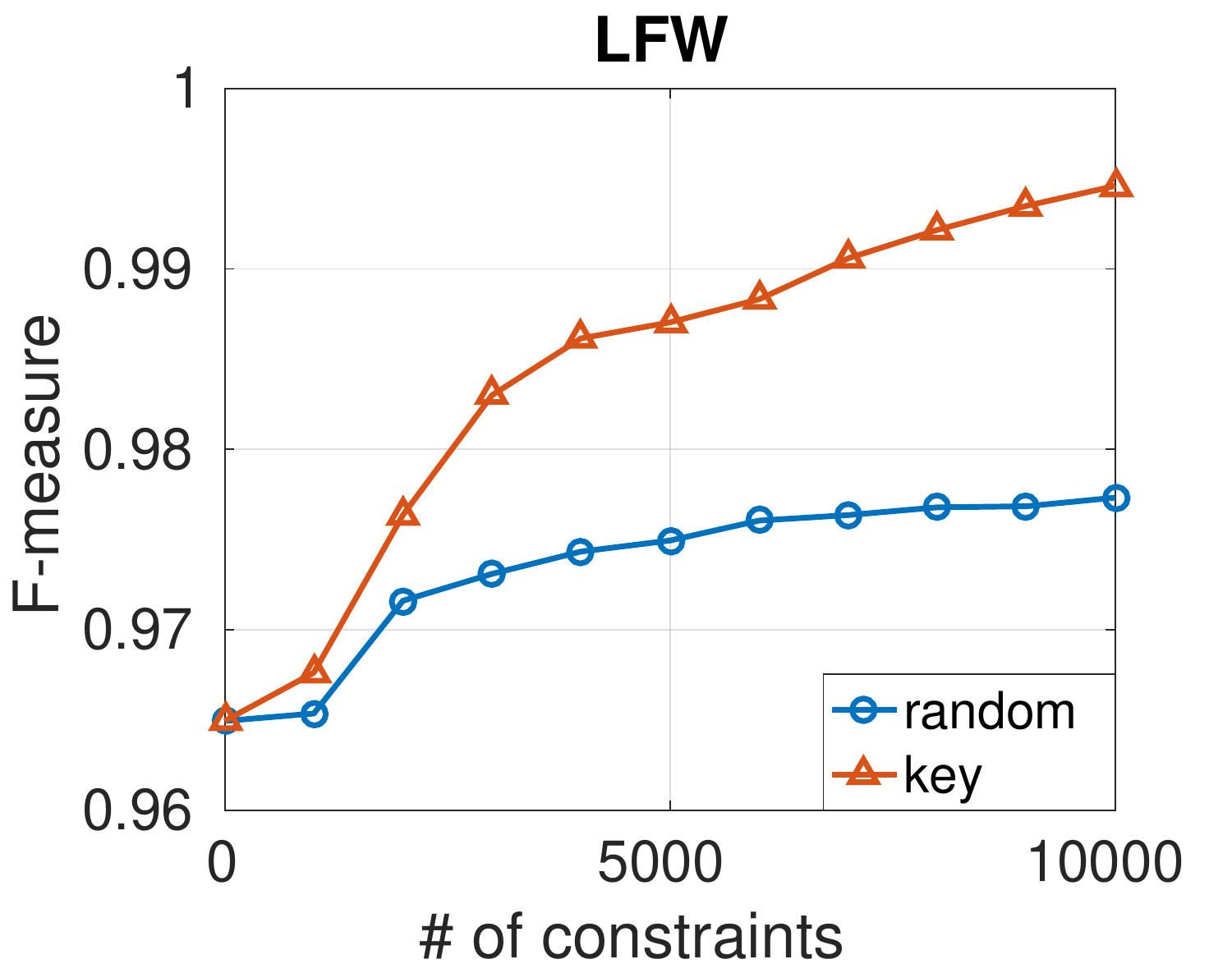}
\includegraphics[width=0.49\columnwidth]{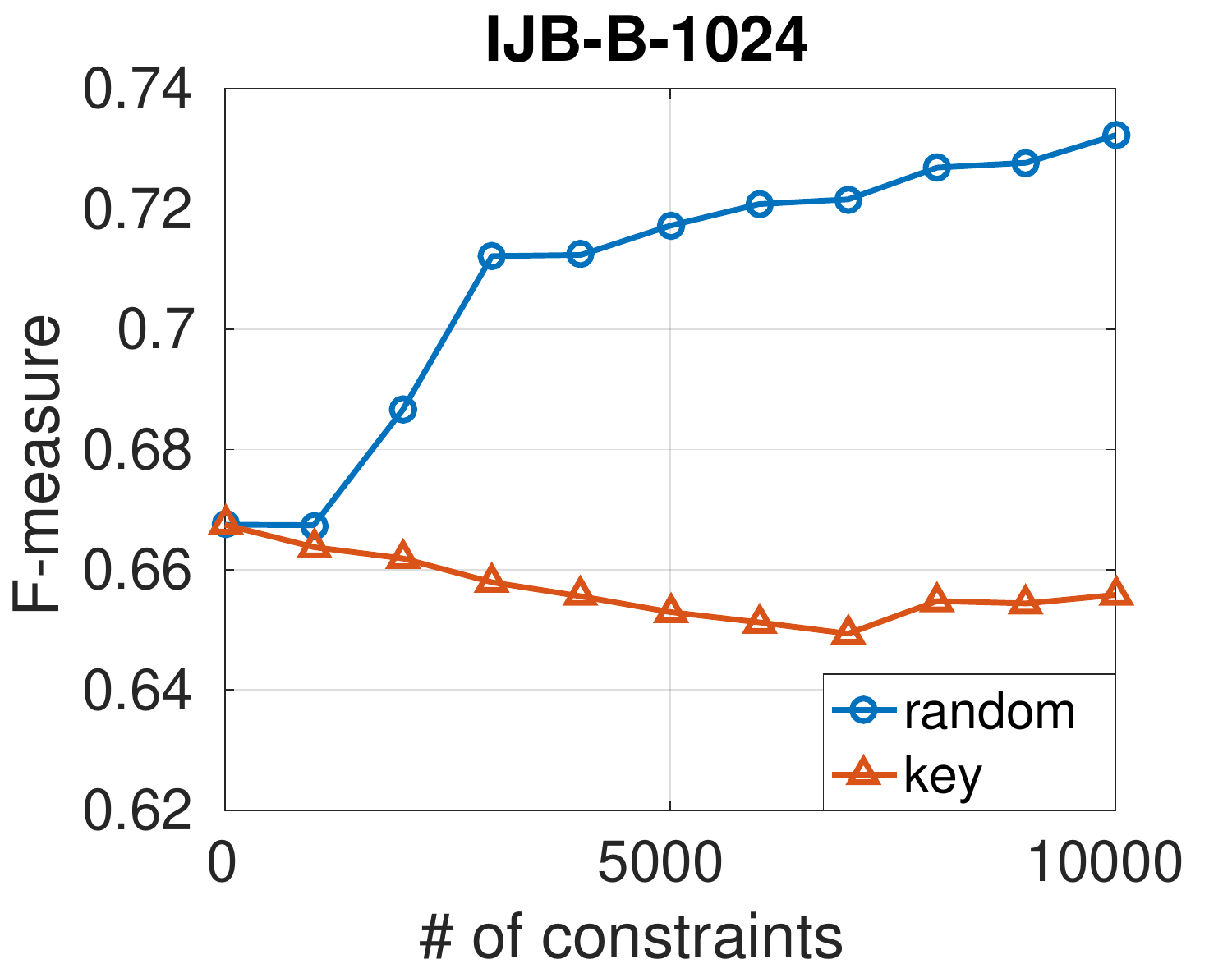}
   \caption{Performance of the proposed clustering algorithm on LFW and IJB-B-1024 datasets after incorporating pairwise constraints. Both random and key constraints improve clustering performance in terms of pairwise F-score on the LFW dataset. However, for the IJB-B-1024 dataset, only random constraints are able to boost the performance.}
\label{fig:score_lfw_constraints}
\end{figure}

\begin{figure}[t]
\subfigure[Must-links]{
\includegraphics[width=\columnwidth]{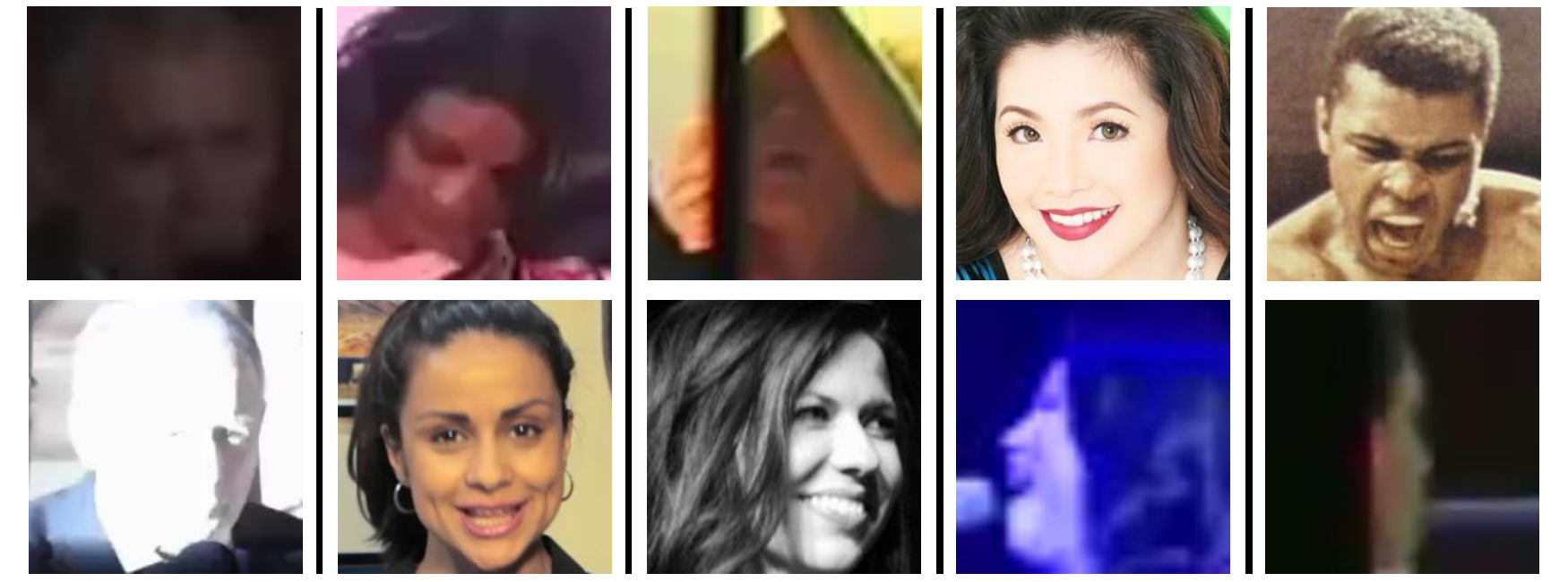}
}
\subfigure[Cannot-links]{
\includegraphics[width=\columnwidth]{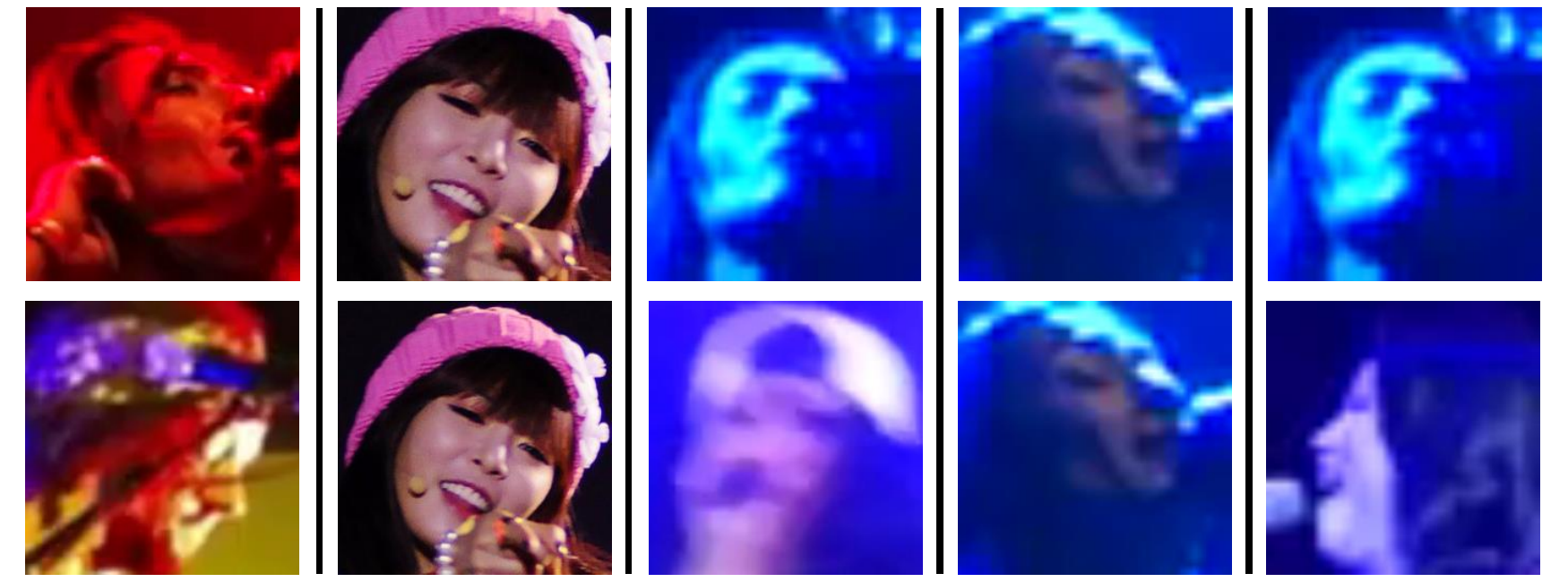}
}
\caption{Example pairs in the key constraints for IJB-B-1024 dataset. Many must-links have extremely different facial appearances, even though they may belong to the same identity. Cannot-links, on contrary, involve some wrong labels and many extremely low-quality images.}
\label{fig:example_key_constraints}
\end{figure}

\begin{table*}[tb]
\caption{Comparison of the F-measures of the proposed algorithms and other clustering algorithm on IJB-B datasets. The number of images in each dataset is given with the dataset name. Numbers of non-singleton clusters in the proposed algorithms are shown in parentheses. ``-" means that algorithm cannot be tested due to memory limit.}
\label{tab:cs3_cluster}
\centering
\scriptsize
\subfigure[IJB-B-32 (1,026)]{
\begin{tabularx}{0.48\linewidth}{Xrrrr}
\toprule
            & Pairwise  & BCubed    & \# of      &           \\
Algorithm   & F-measure & F-measure & Clusters   & Run-time  \\
\midrule
\textit{k}-means     & $0.825$   & $0.766$   & $20$       & $00:00:01$\\
Spectral    & $0.630$   & $0.720$   & $30$       & $00:00:01$\\
SSC         & $0.532$   & $0.665$   & $30$        & $00:00:12$\\
Affinity
Propagation & $0.376$   & $0.534$   & $86$       & $00:00:01$\\
Agglomerative & $0.787$   & $0.795$   & $30$       & $00:00:01$\\
Rank-Order  & $0.697$   & $0.769$   & $81$      & $00:00:03$\\
Approx.
Rank-Order  & $0.223$   & $0.526$   & $85$      & $00:00:02$\\ \hline
ConPaC (proposed)      & $\mathbf{0.868}$   & $\mathbf{0.828}$   & $96$      & $00:00:03$\\
&           &           & $(42)$     &           \\
\bottomrule
\end{tabularx}
}
\subfigure[IJB-B-64 (2,080)]{
\begin{tabularx}{0.48\linewidth}{Xrrrr}
\toprule
            & Pairwise  & BCubed    & \# of      &           \\
Algorithm   & F-measure & F-measure & Clusters   & Run-time  \\
\midrule
\textit{k}-means     & $0.704$   & $0.710$   & $75$       & $00:00:01$\\
Spectral    & $0.564$   & $0.654$   & $50$       & $00:00:02$\\
SSC         & $0.467$   & $0.614$   & $50$       & $00:00:44$\\
Affinity
Propagation & $0.399$   & $0.506$   & $167$      & $00:00:02$\\
Agglomerative & $0.635$   & $0.738$   & $75$       & $00:00:05$\\
Rank-Order  & $0.409$   & $0.693$   & $173$      & $00:00:13$\\
Approx.
Rank-Order  & $0.163$   & $0.543$   & $236$      & $00:00:03$\\ \hline
ConPaC (proposed)      & $\mathbf{0.776}$   & $\mathbf{0.772}$   & $280$      & $00:00:04$\\
&           &           & $(93)$    &           \\
\bottomrule
\end{tabularx}
}\\

\subfigure[IJB-B-128 (5,224)]{
\begin{tabularx}{0.48\linewidth}{Xrrrr}
\toprule
            & Pairwise  & BCubed    & \# of     &           \\
Algorithm   & F-measure & F-measure & Clusters  & Run-time  \\
\midrule
\textit{k}-means    & $0.520$   & $0.655$   & $75$      & $00:00:04$\\
Spectral    & $0.377$   & $0.620$   & $100$      & $00:00:12$\\
SSC         & $0.275$   & $0.547$   & $100$     & $00:05:19$\\
Affinity
Propagation & $0.230$   & $0.441$   & $351$     & $00:00:18$\\
Agglomerative & $0.787$   & $0.750$   & $150$       & $00:00:34$\\
Rank-Order  & $0.385$   & $0.661$   & $471$     & $00:02:01$\\
Approx.
Rank-Order  & $0.302$   & $0.653$   & $631$   & $00:00:05$\\ \hline
ConPaC (proposed)      & $\mathbf{0.895}$   & $\mathbf{0.769}$   & $738$   & $00:00:20$\\
&           &           & $(243)$   &           \\
\bottomrule
\end{tabularx}
}\
\subfigure[IJB-B-256 (9,867)]{
\begin{tabularx}{0.48\linewidth}{Xrrrr}
\toprule
            & Pairwise  & BCubed    & \# of     &           \\
Algorithm   & F-measure & F-measure & Clusters  & Run-time  \\
\midrule
\textit{k}-means     & $0.489$   & $0.624$   & $150$      & $00:00:12$\\
Spectral    & $0.333$   & $0.562$   & $150$     & $00:01:00$\\
SSC         & $0.278$   & $0.498$   & $150$     & $00:26:30$\\
Affinity
Propagation & $0.241$   & $0.434$   & $658$     & $00:01:66$\\
Agglomerative & $0.671$   & $\mathbf{0.725}$   & $350$       & $00:01:57$\\
Rank-Order  & $0.351$   & $0.653$   & $1,033$   & $00:06:37$\\
Approx.
Rank-Order  & $0.214$   & $0.645$   & $13,55$   & $00:00:07$\\ \hline
ConPaC (proposed)      & $\mathbf{0.888}$   & $0.721$   & $1,862$   & $00:00:48$\\
&           &           & $(574)$   &           \\
\bottomrule
\end{tabularx}
}\\

\subfigure[IJB-B-512 (18,251)]{
\begin{tabularx}{0.48\linewidth}{Xrrrr}
\toprule
            & Pairwise  & BCubed    &  \# of    &           \\
Algorithm   & F-measure & F-measure &  Clusters & Run-time  \\
\midrule
\textit{k}-means     & $0.429$   & $0.587$   & $500$     & $00:00:46$\\
Spectral    & $0.335$   & $0.531$   & $350$     & $00:06:28$\\
SSC         & $0.237$   & $0.450$   & $500$     & $02:35:21$\\
Affinity
Propagation & $0.251$   & $0.432$   & $1,276$   & $00:17:16$\\
Agglomerative & $0.567$   & $\mathbf{0.687}$   & $750$       & $00:04:45$\\
Rank-Order  & $0.188$   & $0.638$   & $1,958$   & $00:23:49$\\
Approx.
 Rank-Order & $0.214$   & $0.569$   & $3,758$   & $00:00:12$\\ \hline
ConPaC (proposed)      & $\mathbf{0.756}$   & $0.656$   & $3,981$   & $00:02:10$\\
&           &           & $(1,175)$ &           \\
\bottomrule
\end{tabularx}
}
\subfigure[IJB-B-1024 (36,575)]{
\begin{tabularx}{0.48\linewidth}{Xrrrr}
\toprule
            & Pairwise  & BCubed    &  \# of    &           \\
Algorithm   & F-measure & F-measure &  Clusters & Run-time  \\
\midrule
\textit{k}-means     & $0.423$   & $0.572$   & $1,000$     & $00:03:01$\\
Spectral    & $0.265$   & $0.495$   & $750$     & $00:45:43$\\
SSC         & $-$   & $-$   & $-$   & $-$\\
Affinity
Propagation & $0.241$   & $0.423$   & $2,500$   & $01:37:38$\\
Agglomerative & $0.544$   & $\mathbf{0.696}$   & $1500$       & $00:28:21$\\
Rank-Order  & $0.020$   & $0.544$   & $2,831$   & $01:40:30$\\
Approx.
Rank-Order  & $0.201$   & $0.512$   & $9,553$  & $00:00:23$\\ \hline
ConPaC (proposed)      & $\mathbf{0.667}$  & $0.641$ & $11,258$   & $00:08:39$ \\
&           &           & $(2,303)$ &           \\
\bottomrule
\end{tabularx}
}\\

\subfigure[IJB-B-1845 (68,195)]{
\begin{tabularx}{0.48\linewidth}{Xrrrr}
\toprule
            & Pairwise  & BCubed    &  \# of    &           \\
Algorithm   & F-measure & F-measure &  Clusters & Run-time  \\
\midrule
\textit{k}-means     & $0.354$   & $0.551$   & $1500$    & $00:11:49$\\
Spectral    & $-$   & $-$   & $-$     & $-$\\
SSC         & $-$       & $-$       & $-$       & $-$\\
Affinity
Propagation & $-$       & $-$       & $-$       & $-$\\
Agglomerative & $-$   & $-$   & $-$       & $-$\\
Rank-Order  & $0.005$   & $0.267$   & $4,084$   & $01:12:25$\\
Approx.
Rank-Order  & $0.299$   & $0.450$   & $20,782$  & $00:00:43$\\ \hline
ConPaC (proposed)      & $\mathbf{0.611}$   & $\mathbf{0.634}$   & $15,227$  & $00:51:33$\\
&           &           & $(4,200)$ &           \\
\bottomrule
\end{tabularx}
}
\end{table*}

Another thing worth noticing is that the number of clusters found by the proposed algorithm is much larger than the true number of clusters. This is because a large number of points are regarded as outliers by our algorithm and so they form singleton clusters. For this reason, we also report the number of ``non-singleton" clusters in parentheses, which contain at least two points. The number of non-singleton clusters is closer to the true number of clusters.

Some example clustering results on the IJB-B-32 and IJB-B-1024 are shown in Figure~\ref{fig:example_cs3}. The first two rows show an impure and a pure cluster on IJB-B-32 dataset, respectively. The other two rows show an impure and a pure clsuter on IJB-B-1024 dataset. Many face images in the impure clusters have very large pose variations and are thus badly aligned, diminishing the saliency of the representation.

\subsubsection{Semi-supervised Clustering}

As we mentioned in Section~\ref{sec:semi}, pairwise constraints could be naturally incorporated into the framework of ConPaC without any modification of the algorithm. Therefore in this section, we assume that we have already been given a set of pairwise constraints and evaluate whether the side-information could improve the clustering performance. We consider two types of constraints: 

\begin{itemize}
  \item \textit{Random Constraints}: must-links and cannot-links are picked randomly from ground-truth positive and negative pairs.
  \item \textit{Key Constraints}: The similarities between every pair of faces are sorted. Must-links are picked by choosing the positive pairs (pairs from the same identity) with the lowest similarities and cannot-links are picked by choosing the negative pairs (faces from different identities) with the highest similarities.
\end{itemize}

Key constraints are difficult to acquire in realistic cases, as such they are merely used to test the upper bound of the improvement given constraints on pairs that could be misleading. In both cases, we use knowledge of the ground truth identity labels to sample an equal number of must-link and cannot-link constraints. We then test the performance of the algorithm with an increasing number of constraints. For random constraints, we run $10$ trials and report the average performance. The results are reported in terms of pairwise F-score.


We test our algorithm in a semi-supervised scenario on LFW and IJB-B-1024 datasets. The results of semi-supervised clustering are shown in Figure~\ref{fig:score_lfw_constraints}. On LFW, for both random constraints and key constraints, the constraints always boost the performance and the laerger the number of constraints, the larger the improvement in F-score. This is because our algorithm tries to find clustering results that are most consistent with the unary potentials, and when more constraints are provided, the unary potentials can be trusted more. Additionally, the number of specified constraints in the experiment are actually very small. For example, $10,000$ constraints account for only $0.011\%$ of the total number of all the possible pairs in LFW. But due to message propagation, each pairwise constraint impacts all related pairs. Thus, even a small number of randomly picked constraints could boost the performance significantly. For the $10,000$ random constraints, $98.33\%$ must-links and $99.99\%$ cannot-links are satisfied at the end. For key constraints, $98.42\%$ must-links and $99.94\%$ cannot-links are satisfied. 


On IJB-B-1024 dataset, as it is a larger dataset, $10,000$ constraints account for only $0.0015\%$ of the total number of pairs. The random constraints boost the performance significantly, but the key constraints do not offer any benefits. Furthermore, $99.93\%$ must-links and $75.17\%$ cannot-links are satisfied for random constraints, but only $3.02\%$ must-links and $21.60\%$ cannot-links are satisfied for $10,000$ key constraints. Inspecting into the problem, we found that the selected must-links for key constraints on IJB-B-1024 usually have extremely different facial appearances, while images in cannot-links are mostly of very bad quality, as shown in Figure~\ref{fig:example_key_constraints}. Thus, the algorithm may not satisfy these constraints as they are quite incompatible with the rest of the dataset.

\subsubsection{\textit{k}-NN variant for large datasets}
\begin{figure}[t]
\includegraphics[width=0.49\columnwidth]{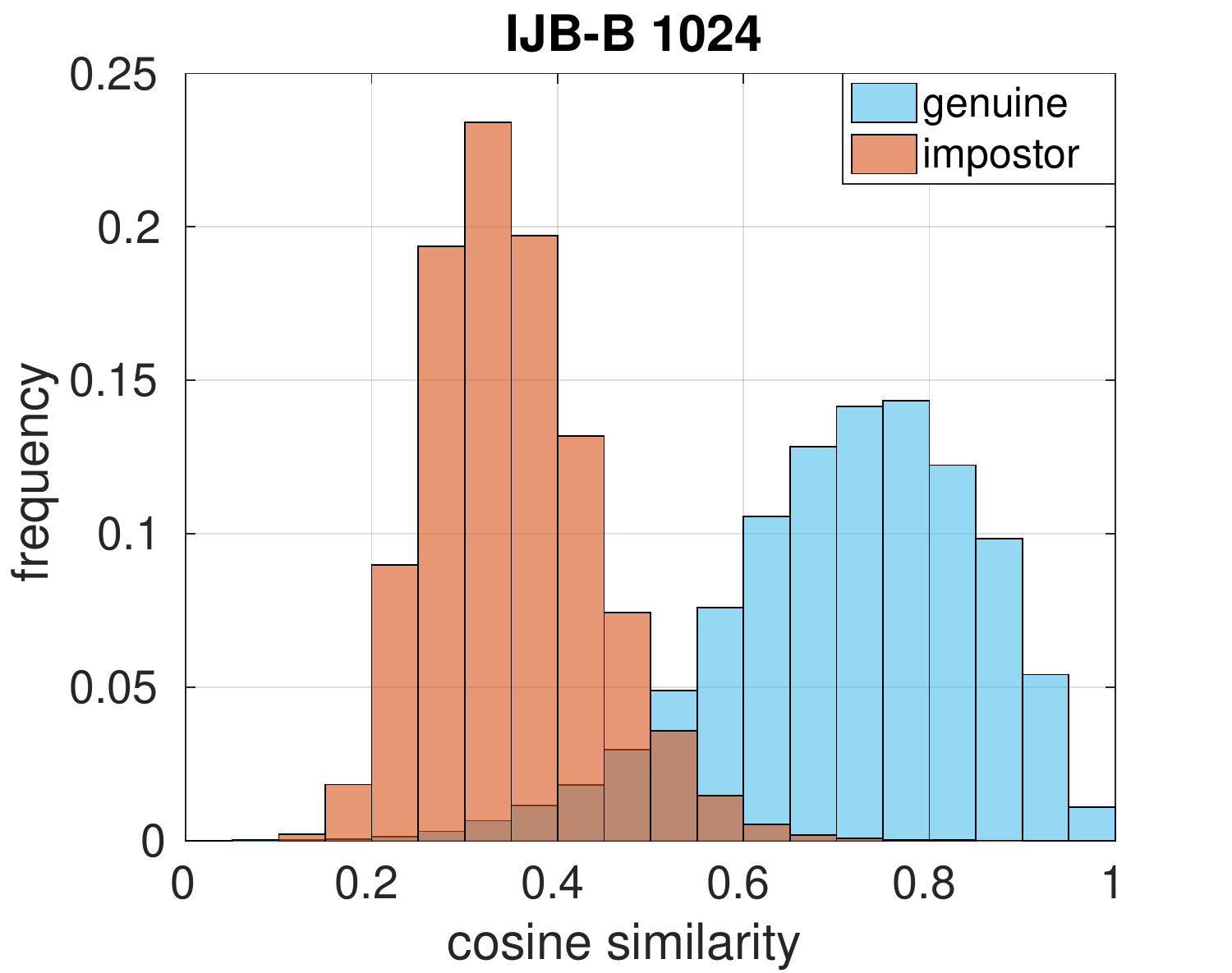}
\includegraphics[width=0.49\columnwidth]{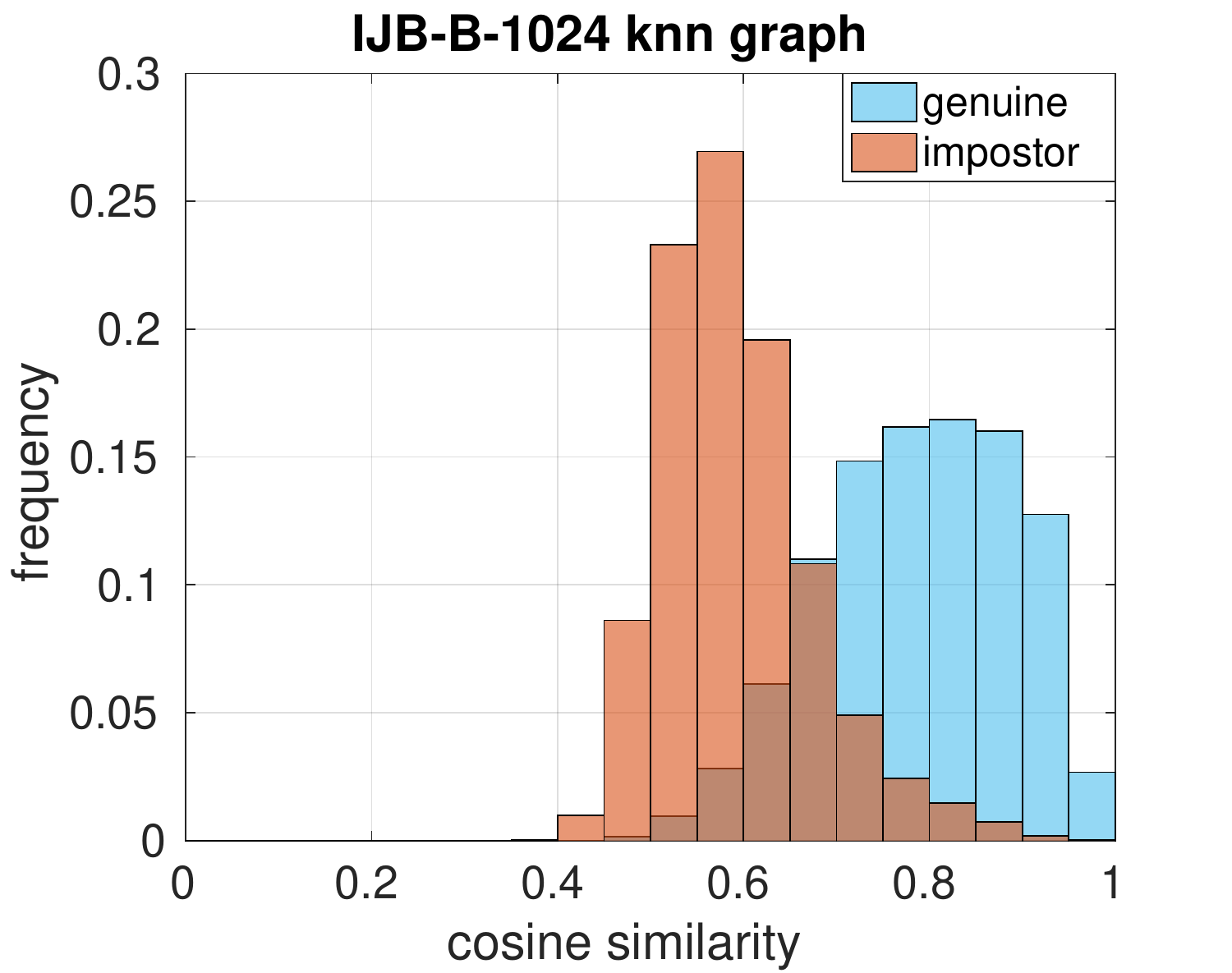}
   \caption{The distributions of all genuine and impostor pairs in IJB-B-1024 and pairs in \textit{k}-NN graph. Both the impostor and genuine distributions are shifted significantly in the k-NN graph.}
\label{fig:score_hist_1024_knn}
\end{figure}

In this section, we test the run-time and performance of the \textit{k}-NN variant of the proposed clustering. We use the same \textit{k}-d tree library~\cite{flann_pami_2014} as used in \cite{otto2017clustering} to generate the approximate \textit{k}-NN graph. We also use the same configuration for the \textit{k}-d tree with $k=200$ and we build four trees with a search size of $2,000$. We first compare the performance of the algorithm using approximate \textit{k}-NN graph and full graph (original algorithm) on LFW and IJB-B-1024. The clustering results are shown in Table~\ref{tab:knn_cluster}. The proposed \textit{k}-NN variant performs well on both datasets. For LFW, the pairwise F-measure is almost as good as the original algorithm. For IJB-B-1024, we observe a large shift of the distributions of similarities in the \textit{k}-NN graph compared with original distributions, especially for impostor pairs, as shown in Figure~\ref{fig:score_hist_1024_knn}. This is because the \textit{k}-d tree is expected to select only similar pairs for building the \textit{k}-NN graph, and many impostor pairs in IJB-B-1024 dataset turn out to be similar. Using old threshold $\tau=0.7$ would make the majority of pairs positive, thus we tune the threshold to $\tau=0.75$ for \textit{k}-NN variant when working on IJB-B-1024. Although a similar shift is observed on LFW dataset, but it is smaller and we found that there is no need for tuning the threshold. After tuning the threshold, the F-measure of \textit{k}-NN variant is also close to the original one on IJB-B-1024. Using the same threshld $\tau$, we then test the performance of the \textit{k}-NN variant on $1$ million unlabeled face images along with LFW or IJB-B. The 1 million face images is a subset of the same private dataset used in~\cite{otto2017clustering}. Since we do not have the labels for the $1$ million dataset, we apply pairwise F-measure to only the subset for which we have labels (from LFW or IJB-B) but omit those for which we do not have labels, namely the $1$ million distractor images. With such a big number of distractors, the resulting F-measure is almost unchanged for both datasets. Notice that for IJB-B-1024, the performance is, surprisingly, even better when given the distractors. A plausible explanation for this is that there are some high quality images in the 1 million distractors that are from the same identities as in IJB-B, which help to reveal the connections between the face images in IJB-1024, while the other distractor images make less impact on the clustering performance.


\begin{table}[t]
\label{tab:knn}
\centering
\caption{Performance of \textit{k}-NN variant and original algorithm on small and large datasets.}
\begin{tabularx}{\columnwidth}{Xrrrr}
\toprule
            &             & Pairwise  &  \# of     &           \\
Dataset     & Version     & F-measure & clusters   & Run-time  \\
\midrule
LFW         & full graph            & $0.965$   & $6,352$    & $00:00:39$\\
LFW         & \textit{k}-d tree     & $0.960$   & $6,444$    & $00:01:04$\\
IJB-B-1024  & full graph            & $0.668$   & $8,094$    & $00:08:39$\\
IJB-B-1024  & \textit{k}-d tree     & $0.536$   & $11,258$   & $00:01:43$\\
LFW + 1M    & \textit{k}-d tree     & $0.940$   & $536,809$  & $00:34:03$\\
IJB-B-1024 + 
1M  & \textit{k}-d tree     & $0.541$   & $616,964$  & $00:36:05$\\
\bottomrule
\end{tabularx}
\label{tab:knn_cluster}
\end{table}

\subsubsection{Influence of the initial similarity matrix}
\label{sec:quality_sim}

The motivation and distinguishing feature of our algorithm is that it only depends on the given pairwise similarities. In this subsection we want to investigate how the clustering performance is affected by the similarities and also how it is influenced by the choice of parameter $\tau$.

We first define \textit{similarity reliability} as the pairwise F-measure on adjacency matrix $Z$, where $Z_{ij}\in\{0,1\}$ is determined by thresholding the cosine similarity between $X_i$ and $X_j$ by $\tau$. Different from $Y$, the graph represented by $Z$ may not be transitive, so it does not necessarily correspond to a clustering result.

\begin{figure}[t]
\center
\subfigure[]{
\includegraphics[width=0.465\columnwidth]{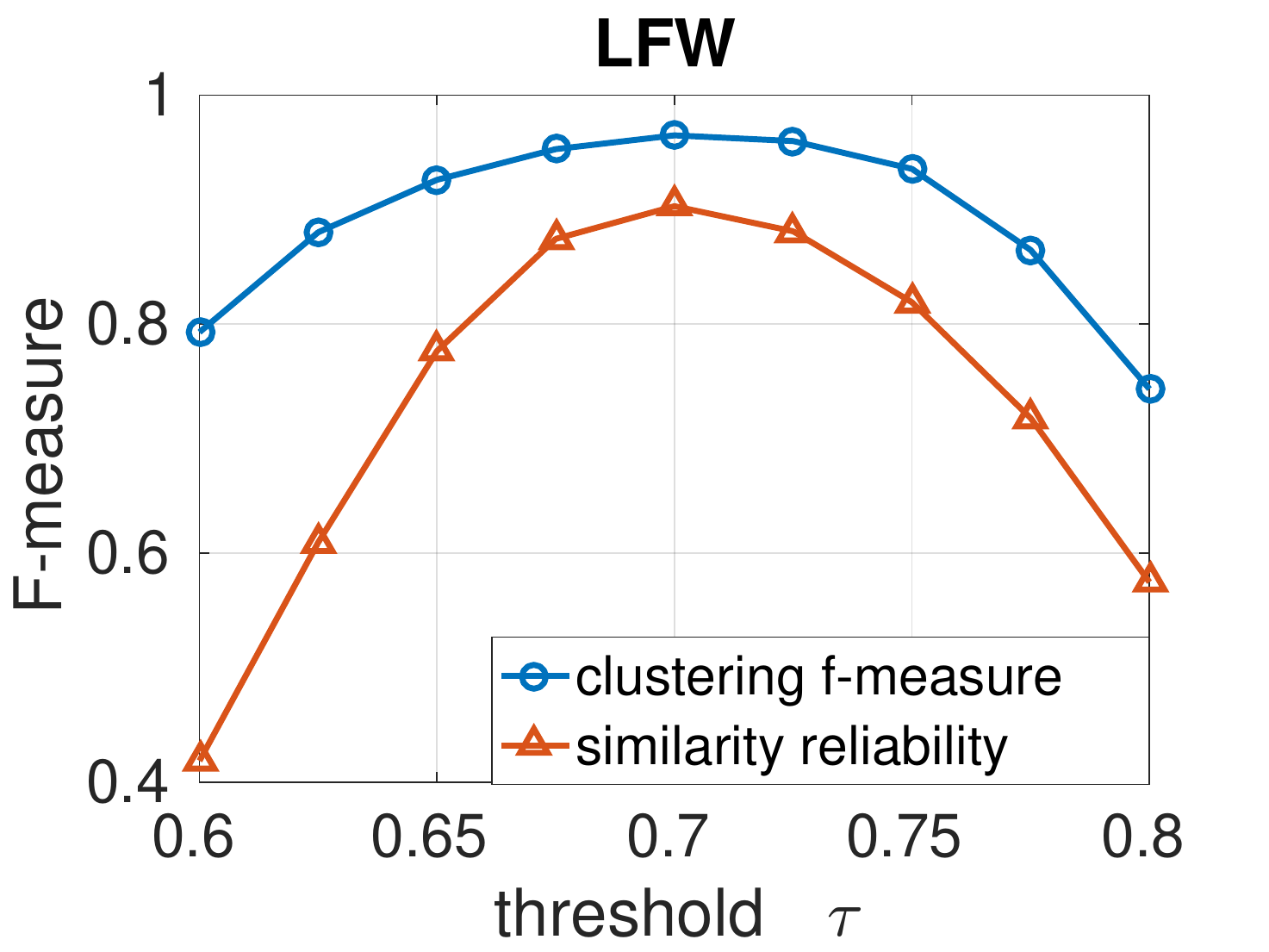}
}
\subfigure[]{
\includegraphics[width=0.465\columnwidth]{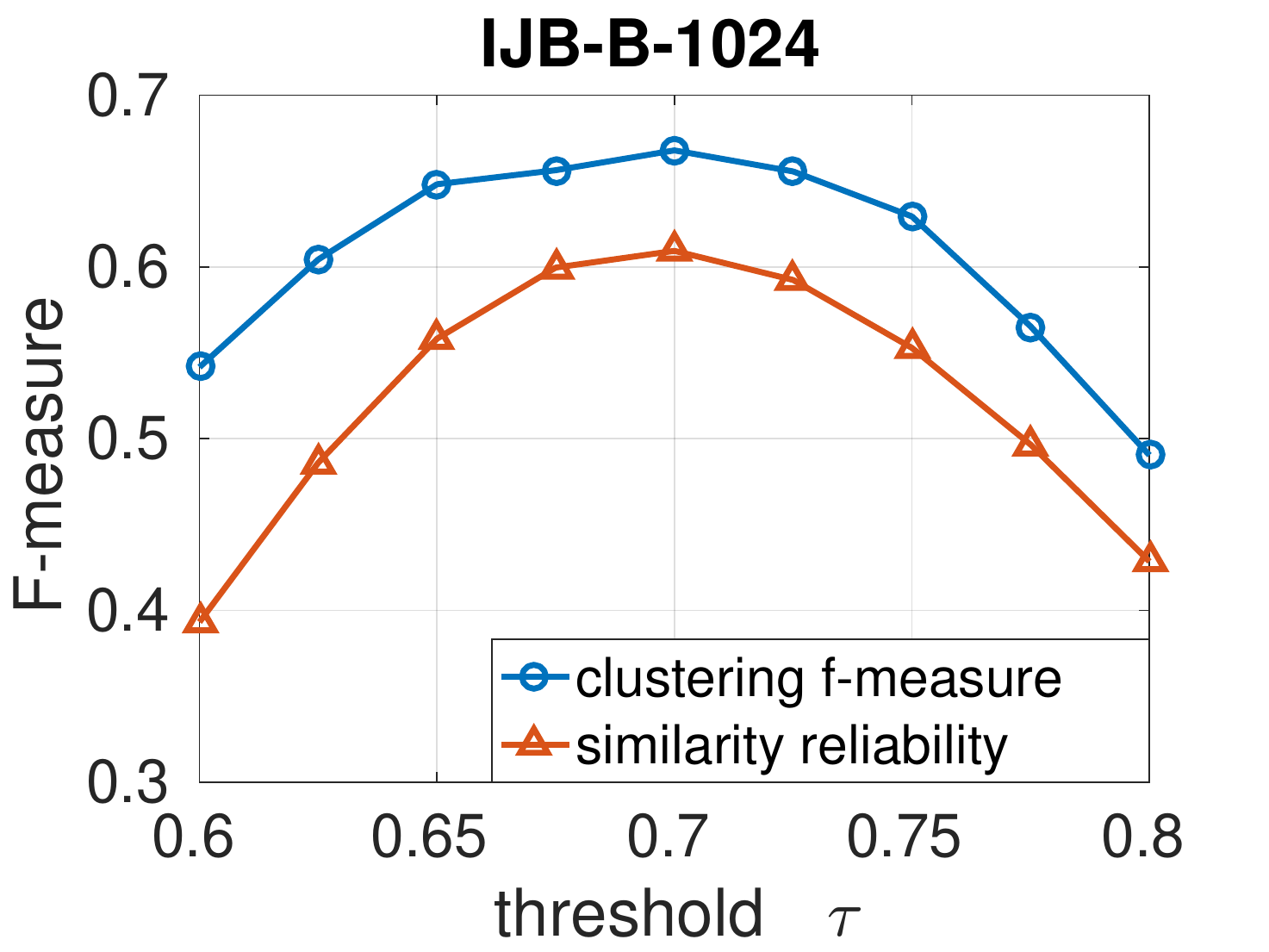}
}
 \caption{Performance of the proposed algorithm with different threshold values on LFW and IJB-B-1024 datasets.}
\label{fig:score_thresholds}
\end{figure}

\begin{figure}[t]
\center
\subfigure[]{
\includegraphics[width=0.465\columnwidth]{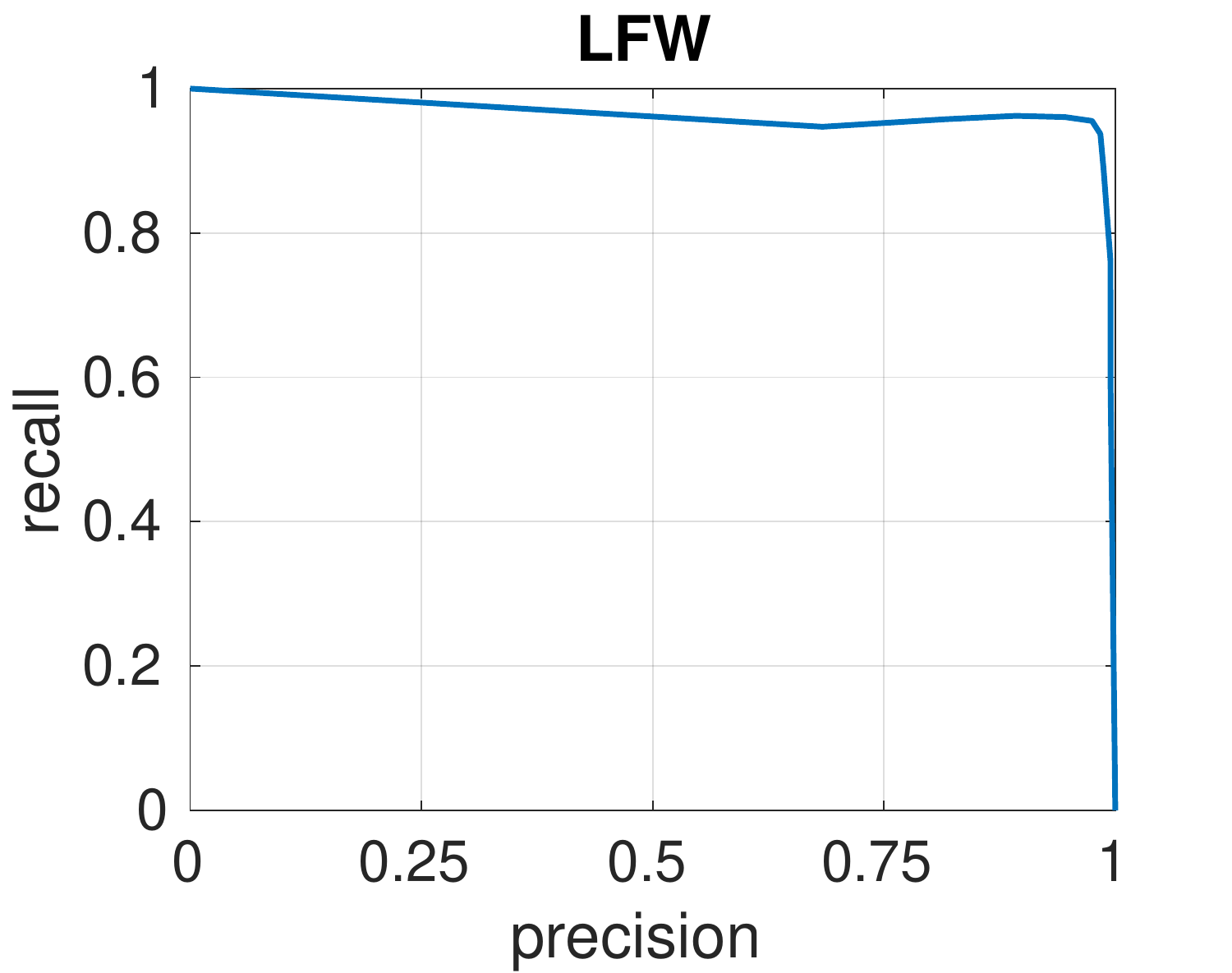}
}
\subfigure[]{
\includegraphics[width=0.465\columnwidth]{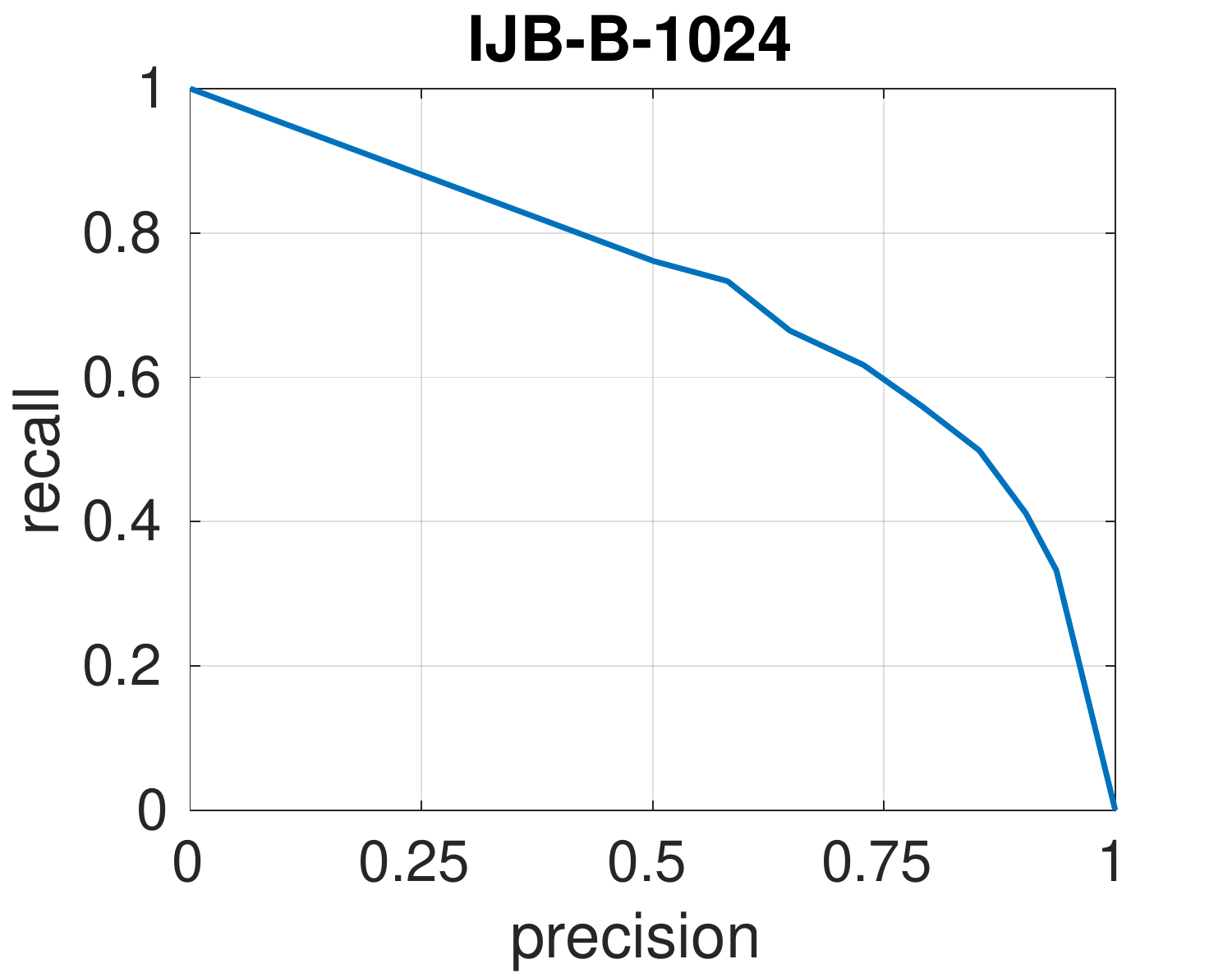}
}
 \caption{Precision Recall curves for the proposed algorithm on LFW and IJB-B-1024 datasets.}
\label{fig:pr_curve}
\end{figure}

We then determine how the F-measures of $Y$ and $Z$ vary with different values of threshold $\tau$. The results are shown in Figure~\ref{fig:score_thresholds}, and corresponding precision-recall curves are shown in Figure~\ref{fig:pr_curve}. We can see that the clustering performance changes smoothly with different parameter values. Furthermore, the F-measure of clustering result $Y$ is highly correlated with that of $Z$. In other words, when the similarity matrix is reliable, the clustering does a better job, and vice versa.


To further see the relationship between the clustering performance and the pairwise similarities, we compare the F-measures of $Y$ and that of $Z$ on all the 7 experiments in IJB-B dataset. We find that the two F-measures are almost linearly correlated across all these experiments, with a correlation coefficient of $0.9998$. Therefore, we can state that we have achieved our motivation to make full use of the input pairwise similarities, and that the decrease in clustering performance on IJB-B compared to LFW is due to the decrease in reliability of the input pairwise similarities, which in turn depends on the saliency of the face representation (feature vector).

\section{Conclusions}

In this paper, we first trained a ResNet deep network architecture on CASIA-Webface and VGG-Face datasets. The representation from the proposed network shows a good performance on the BLUFR face verification benchmark. Using this representation, we proposed a new clustering algorithm, called Conditional Pairwise Clustering (ConPaC), which learns an adjacency matrix directly from the given similarity matrix. The clustering problem is modeled as a structured prediction problem using a Conditional Random Field (CRF) and is inferred by Loopy Belief Propagation. The proposed algorithm outperforms several well known clustering algorithms on LFW and IJB-B unconstrained datasets and it can also naturally incorporate pairwise constraints to further improve clustering results. We also propose a \textit{k}-NN variant of ConPaC which is capable of clustering millions of face images. Our future work would include finding better unary potentials for more robust face clustering and also incorporating pairwise constraints into the \textit{k}-NN variant.


{\small
\bibliographystyle{IEEEtran}
\bibliography{ref}
}
%

\newcommand{\biospace}{\vspace{-0.5in}}
\biospace
\begin{IEEEbiography}[{\includegraphics[width=1in,height=1.25in,clip,keepaspectratio]{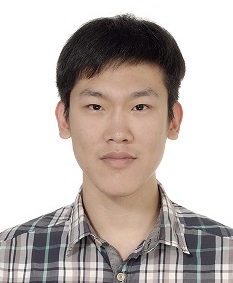}}]{Yichun~Shi} received his B.S degree in the Department of Computer Science and Engineering at Shanghai Jiao Tong University in 2016. He is now working towards the Ph.D. degree in the Department of Computer Science and Engineering at Michigan State University. His research interests include pattern recognition and computer vision.
\end{IEEEbiography}

\biospace
\begin{IEEEbiography}[{\includegraphics[width=1in,height=1.25in,clip,keepaspectratio]{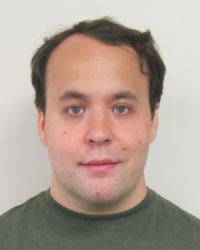}}]{Charles~Otto} received his B.S. and Ph.D. degrees in the Department of Computer Science and Engineering at Michigan State University in 2008 and 2016, respectively. He was a research engineer at IBM during 2006-2011.  He is currently employed at Noblis, Reston, VA. His research interests include pattern recognition, and computer vision.
\end{IEEEbiography}

\biospace
\begin{IEEEbiography}[{\includegraphics[width=1in,height=1.25in,clip,keepaspectratio]{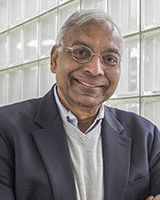}}]{Anil~K.~Jain} is a University distinguished professor in the Department of Computer Science and Engineering at Michigan State University. His research interests include pattern recognition and biometric authentication. He served as the editor-in-chief of the IEEE Transactions on Pattern Analysis and Machine Intelligence (1991-1994), a member of the United States Defense Science Board, and and is a member of the National Academy of Engineering.
\end{IEEEbiography}




\end{document}